\documentclass[lettersize,journal]{IEEEtran}

\usepackage{amsmath,amsfonts,amssymb}
\usepackage{algorithm}
\usepackage{array}
\usepackage[caption=false,font=normalsize,labelfont=sf,textfont=sf]{subfig}
\usepackage{textcomp}
\usepackage{stfloats}
\usepackage{url}
\usepackage{verbatim}
\usepackage{graphicx}
\usepackage{booktabs}
\usepackage{multirow}
\usepackage{cite}
\usepackage{xcolor}
\usepackage{algpseudocode}
\begin{document}

\title{NutriVision: Ingredient-Conditioned Fusion and Prediction for Single-Image Food Nutrition Estimation}

\author{
Avinash~Anand,
Aman~Kumar,
Chaitanya~Lakhchaura,
Ashutosh~Kumar,
Akshita~Abrol,
Timothy~Liu,
and Zhengkui~Wang
\thanks{Avinash Anand, Chaitanya Lakhchaura, Akshita Abrol, and Zhengkui Wang are with the
Singapore Institute of Technology, Singapore (e-mail: Avinash.Anand@singaporetech.edu.sg; Chaitanya.Lakhchaura@Singaporetech.edu.sg; akshita.abrol@singaporetech.edu.sg; zhengkui.wang@singaporetech.edu.sg).}%
\thanks{Aman Kumar is with the
Indraprastha Institute of Information Technology Delhi (IIIT-Delhi),
New Delhi, India
(e-mail: aman24012@iiitd.ac.in).}%

\thanks{Ashutosh Kumar is with the Rochester Institute of Technology,
Rochester, NY, USA
(e-mail: ak1825@rit.edu)}%
\thanks{Timothy Liu is with the NVIDIA AI Technology Centre, Singapore
(e-mail: timothyl@nvidia.com).}%

}



\maketitle

\begin{abstract}

Nutrition estimation is a fundamental task in consumer diet tracking, clinical dietetics, chronic disease management, sports and hospital nutrition, and broader food computing systems. The existing approaches have progressed along two largely separate axes, vision models that rely on calibrated RGB-depth captures and ingredient-aware methods that use textual cues but use limited multimodal fusion. We introduce NutriVision, an end-to-end framework that leverages visual geometry and ingredient semantics to estimate calories, mass, fat content, carbohydrates, and protein from a single RGB image and an optional ingredient list. It obtains the unavailable depth modality using DepthAnything-V3 and encodes ingredient descriptions using CLIP. It integrates three complementary mechanisms: (1) an \emph{Ingredient-Conditioned Frequency-Aligned Fusion Module (IC-FAFM)}, which uses textual guidance to reweight and align RGB-depth frequency components; (2) an \emph{Ingredient-Aware Mask-based Prediction Head (IA-MPH)}, whose gating and channel masks are conditioned on food identity; and (3) modality-specific \emph{Internal Semantic Modeling (ISM)} blocks. On the Nutrition5k dataset, NutriVision achieves a mean PMAE of $\mathbf{13.60\pm0.10\%}$, outperforming our IGSMNet implementation by $0.89$ percentage points and OmniFood8k by $2.90$ percentage points (both $p<0.001$). The module-level ablations identify the ingredient-aware prediction head as the primary architectural contributor, improving mean PMAE by $1.50\pm0.17$ percentage points ($p<0.001$). These results demonstrate that ingredient-conditioned prediction and frequency-aware RGB-depth fusion provide measurable gains for single-image nutrient estimation. More broadly, NutriVision offers a practical route toward nutrition-assessment systems that exploit geometric and semantic cues without requiring specialized depth-sensing hardware.
\end{abstract}

\begin{IEEEkeywords}
Nutrition estimation, multimodal fusion, food image analysis, monocular depth estimation.
\end{IEEEkeywords}

\section{Introduction}
\IEEEPARstart{D}{ietary} assessment is foundational to nutrition science, chronic-disease management, clinical dietetics, and consumer health, while food computing has emerged as a distinct research area over the
past decade~\cite{min2019survey}. Conventional dietary assessment relies heavily on manual food logging, which is inherently time-consuming and susceptible to reporting errors. These limitations motivate \emph{image-based food nutrition estimation}, the prediction of a meal's energy, mass, fat, carbohydrate, and protein content directly from a food image~\cite{thames2021nutrition5k,yu2026omnifood8k,foods14213697}. The earlier works on food computing primarily addressed food recognition, exemplified by datasets such as Food-101~\cite{bossard2014food}. Nutrition estimation is substantially more challenging because it requires jointly reasoning about food identity, ingredient composition, and portion size. The release of \textbf{Nutrition5k}~\cite{thames2021nutrition5k}, a large-scale dataset containing ingredient-level nutritional annotations and calibrated RGB-D captures, has
enabled systematic evaluation of models for this task.

Recent methods have improved nutrition estimation through two complementary sources of information: geometric cues derived from depth and semantic cues derived from ingredient descriptions. Geometry-oriented approaches seek to recover the spatial and volumetric information that is difficult to infer from RGB modality alone. For example, \textbf{OmniFood8k}~\cite{yu2026omnifood8k} predicts depth from a single RGB image and processes the RGB and estimated-depth streams using separate encoders. Its Scale-Shift Residual Adapter (SSRA) refines the predicted depth, while its Frequency-Aligned Fusion Module (FAFM) hierarchically aligns the two modalities in the Fourier domain. A Mask-based Prediction Head (MPH) then dynamically selects informative feature channels, yielding a mean PMAE of $16.5\%$ on Nutrition5k.

Ingredient-aware approaches instead exploit explicit information about food composition. This direction builds on earlier work connecting food images with recipes and ingredient descriptions, such as Recipe1M~\cite{salvador2017learning}. More recently, \textbf{IGSMNet}~\cite{foods14213697} encodes ingredient lists using CLIP~\cite{radford2021learning} and employs cross-attention to align the resulting textual representation with hierarchically fused RGB-D features. Its Internal Semantic Modelling (ISM) mechanism further captures localized spatial and semantic relationships through fine-grained attention and dynamic positional encoding. By explicitly modelling ingredient semantics, IGSMNet obtains a mean PMAE of $15.0\%$ on Nutrition5k, with particularly strong improvements for identity-sensitive nutritional quantities such as fat, carbohydrate, and protein.

Despite this progress, geometric and ingredient-level information remain only partially integrated. Frequency-domain RGB-depth fusion captures complementary appearance and structural cues, but its alignment is independent of the ingredients present in the dish. Conversely, ingredient-aware approaches introduce semantic information after conventional visual fusion and therefore do not allow food composition to directly regulate cross-modal interaction or nutrient-specific feature selection. Existing methods also either depend on calibrated depth sensors or underexploit estimated depth as a semantically conditioned geometric representation. This motivates a unified formulation in which ingredient semantics guide both RGB-depth fusion and final nutrient prediction while retaining single-image deployability through monocular depth estimation.
\begin{figure*}
    \centering
    \includegraphics[width=\linewidth]{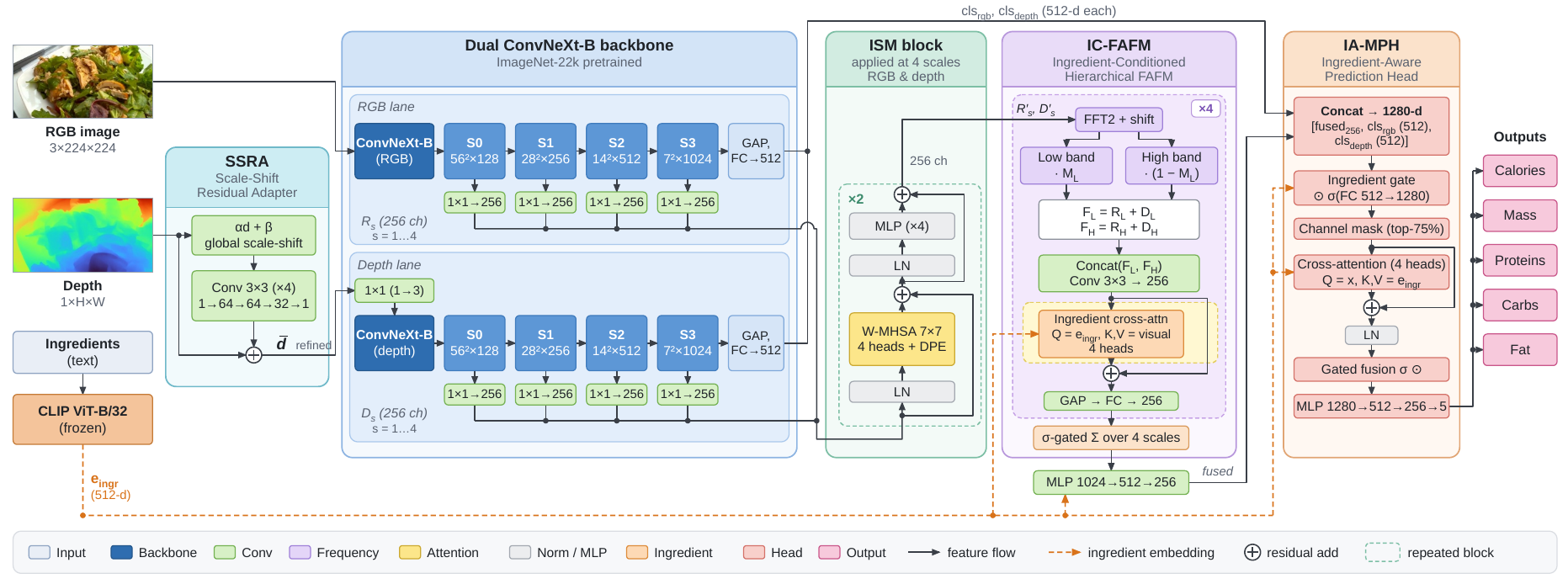}
    \caption{Overview of the proposed ingredient-aware RGB–depth fusion architecture for food nutrition estimation. A scale-shift residual adapter (SSRA) calibrates the depth input, while dual ConvNeXt-B backbones extract multi-scale RGB and depth features. An Ingredient-aware Spatial Mixer (ISM) captures spatial dependencies, followed by an Ingredient-Conditioned Hierarchical Feature Attention Module (IC-FAFM) that integrates ingredient semantics with multi-scale visual features. An Ingredient-Aware Prediction Head (IA-MPH) further refines the fused representation through ingredient-guided cross-attention to jointly predict calories, mass, protein, carbohydrates, and fat.}
    \label{fig:pipeline}
\end{figure*}

\subsection*{Contributions}

\begin{figure}[!t]
\centering
\includegraphics[width=\columnwidth]{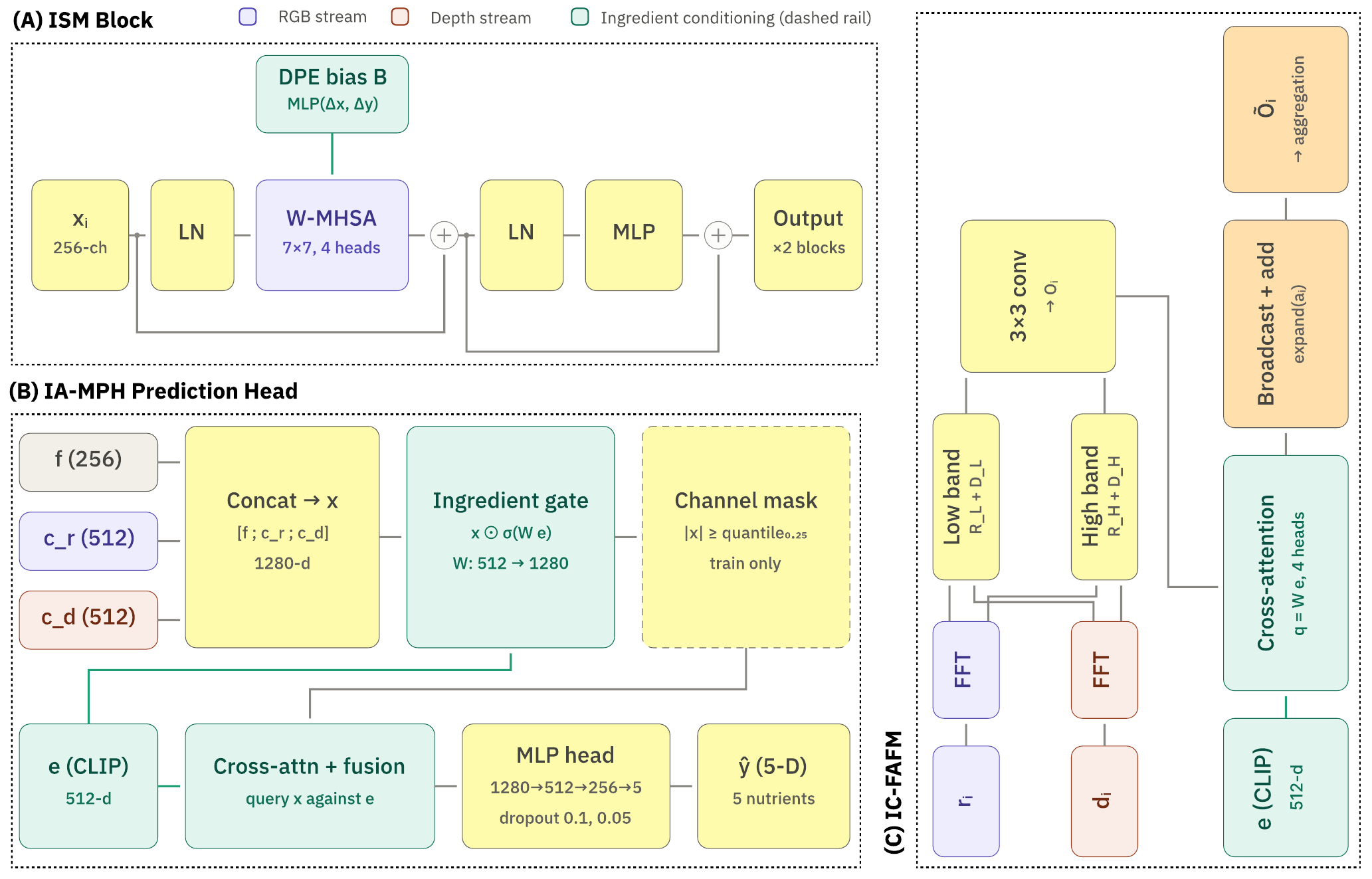}
\caption{Detailed structure of the three NutriVision modules. (A) The ISM block: windowed multi-head self-attention (7×7, 4 heads) with a DPE-derived positional bias $\text{B} = \text{MLP}(\Delta x, \Delta y)$ in a pre-norm residual layout ($\oplus$), applied twice per scale independently to each modality. (B) IC-FAFM at one scale: per-modality 2-D FFTs are split at the low-pass radius $\tau \cdot r_{max} (\tau = 0.20)$, the low and high bands are summed across modalities and merged by a 3×3 convolution into $O_i$, which is then modulated by ingredient cross-attention (q = We) with a broadcast residual to yield $\tilde{O}_i$. (C) IA-MPH: the concatenation $x = [f; c_r; c_d] \in \mathbb{R}^{1280}$ is gated by $\sigma(We)$, masked to the top-75\% channels by magnitude during training only (dashed), and regressed via a second ingredient cross-attention and a three-layer MLP $(1280 \rightarrow 512 \rightarrow 256 \rightarrow 5)$}
\label{fig:nutrivision_components}
\end{figure}

We propose \textbf{NutriVision} (the overall pipeline in Figure~\ref{fig:pipeline} and component details in Figure~\ref{fig:nutrivision_components}), a unified framework that estimates calories, mass, fat, carbohydrates, and protein from an RGB image, an estimated depth map, and an optional ingredient list. Its principal contributions are:

\begin{itemize}

\item \textbf{Ingredient-Conditioned Frequency-Aligned Fusion Module
(IC-FAFM).}
We extend OmniFood8k's FAFM to perform ingredient-conditioned RGB-depth fusion across four ConvNeXt feature scales. At each scale, FFT-based low- and high-frequency components are modulated using a CLIP ingredient representation through multi-head cross-attention, allowing food composition to influence cross-modal spectral alignment.

\item \textbf{Ingredient-Aware Mask-based Prediction Head (IA-MPH).}
We introduce an ingredient-conditioned regression head comprising a semantic gate, magnitude-based top-$k$ channel selection, and ingredient-to-visual cross-attention. This design allows the channels used for nutrient prediction to vary according to the semantic composition of the dish rather than being determined solely by visual activations.

\item \textbf{Modality-specific Internal Semantic Modeling (ISM).}
We extend IGSMNet's ISM mechanism to a dual ConvNeXt-base architecture by applying DPE- and FGM-based ISM blocks independently to the RGB and estimated-depth feature hierarchies before fusion. This enables each modality to develop localized semantic representations while preserving modality-specific information.

\item \textbf{Statistically validated performance.}
Under a three-seed retraining protocol, NutriVision achieves a mean PMAE of
$\mathbf{13.60\pm0.10\%}$ on Nutrition5k, improving upon our IGSMNet reimplementation by $0.89$ percentage points and OmniFood8k by $2.90$ percentage points. Module-level retraining ablations identify IA-MPH as the largest individual architectural contributor, increasing PMAE by $1.50\pm0.17$ percentage points when removed.

\item \textbf{Comprehensive modality and module ablations.}
We conduct test-time input ablations to quantify the contribution of RGB, estimated depth, and ingredient information, together with module-level retraining experiments that isolate the effects of IC-FAFM, IA-MPH, modality-specific ISM, and the training objective.

\end{itemize}

NutriVision targets settings in which ingredient information is available from the user or the serving environment, including diet-tracking applications, institutional canteens, and hospital nutrition systems. Fully autonomous image-only estimation would additionally require an upstream ingredient-recognition component and is outside the scope of this work.


\section{Related Work}

\subsection{Image-Based Food Nutrition Regression}

The Nutrition5k benchmark~\cite{thames2021nutrition5k} established the direct regression of nutritional quantities from food images, defining the task as RGB or RGB-D input $\rightarrow$ calories, mass, fat, carbohydrates, and protein. Its Inception-based baseline achieved a mean PMAE of $18.9\%$ and demonstrated that visual appearance alone is insufficient for reliably estimating both food composition and portion size. Subsequent methods improved feature extraction using stronger convolutional and attention-based architectures. Swin-Nutrition~\cite{foods11213429}, for example, employed a hierarchical Swin Transformer~\cite{liu2021swin} and achieved a mean PMAE of $17.2\%$ using RGB images alone. Later methods increasingly incorporated depth and ingredient information to provide geometric and semantic cues that are difficult to recover from RGB appearance alone.

\subsection{Depth-Assisted Nutrition Estimation}

Depth information provides geometric cues related to food height, volume, and spatial arrangement, making it particularly useful for portion and mass estimation. Early RGB-D methods used the calibrated depth captures supplied by Nutrition5k. FBFPN~\cite{ma2025fbfpn} introduced bidirectional multi-scale RGB-D feature fusion and reported a mean PMAE of $17.3\%$. NuNet, proposed by Kwan~\textit{et al.}~\cite{kwan2025nutrition}, employed parallel
multi-scale transformer branches and dedicated RGB-D fusion modules, achieving $15.65\%$ using real sensor depth. These approaches demonstrate the value of geometric information but assume access to calibrated RGB-D acquisition hardware.

To reduce this hardware dependence, estimated-depth methods synthesize a depth map from a single RGB image. DPF-Nutrition~\cite{foods12234293} introduced an end-to-end depth-prediction and RGB-depth fusion framework, achieving a mean PMAE of approximately $17.8\%$ on Nutrition5k. OmniFood8k~\cite{yu2026omnifood8k} further combined monocular depth estimation with a Scale-Shift Residual Adapter (SSRA), which corrects the global scale and local structure of the predicted depth representation. NutriVision follows this hardware-independent setting, using DepthAnything-V3~\cite{lin2025depth} to generate depth from the input RGB image and SSRA to adapt the resulting relative geometry for nutrition estimation.

\subsection{Frequency-Domain RGB-Depth Fusion}

Frequency-domain processing provides an alternative to conventional spatial-domain multimodal fusion. Low-frequency components primarily encode coarse structure and global layout, whereas high-frequency components preserve
boundaries, textures, and local details. General-purpose architectures such as GFNet~\cite{NEURIPS2021_07e87c2f} and Fast Fourier Convolution~\cite{NEURIPS2020_2fd5d41e} have demonstrated that Fourier-domain operations can efficiently model global dependencies and complementary spectral information.

Within food nutrition estimation, OmniFood8k~\cite{yu2026omnifood8k} introduced a Frequency-Aligned Fusion Module (FAFM) that hierarchically aligns and fuses RGB and adapted depth features in the frequency domain. Its Mask-based
prediction head subsequently performs dynamic visual channel selection for nutrient regression. However, both frequency alignment and channel selection are determined from visual features alone. NutriVision builds on this formulation by conditioning frequency-domain RGB-depth interaction and final channel selection on ingredient semantics.

\subsection{Ingredient-Guided Nutrition Estimation}

Ingredient descriptions provide direct semantic evidence about food composition and therefore offer strong priors for estimating identity-sensitive quantities such as fat, carbohydrates, and protein. Earlier multimodal food research, including Recipe1M~\cite{salvador2017learning}, established the value of aligning food images with recipes and ingredient text. More recent nutrition-estimation methods incorporate these semantics directly into regression architectures.

IMIR-Net~\cite{NIAN2024104664} introduced ingredient-guided multimodal interaction and refinement within an RGB-D framework, reporting a mean PMAE of $17.4\%$. IGSMNet~\cite{foods14213697} subsequently employed an asymmetric dual-branch Swin-Tiny architecture to extract hierarchical RGB and depth representations. After fusing the two visual streams across scales, it uses CLIP~\cite{radford2021learning} ingredient embeddings as queries in a cross-attention
module that guides the fused visual features. Dynamic positional encoding and fine-grained modeling are then applied through its ISM module to capture localized spatial and compositional relationships. IGSMNet reports a mean PMAE of $15.0\%$ on Nutrition5k.

NutriVision adopts CLIP-based ingredient encoding but changes where and how ingredient semantics enter the architecture. Rather than applying ingredient guidance only after hierarchical RGB-D fusion, it conditions the cross-modal frequency-alignment process itself. It additionally uses ingredient information within the prediction head and applies modality-specific semantic modeling to the RGB and estimated-depth streams before fusion. Thus, ingredient information acts as a persistent conditioning signal for representation refinement, cross-modal interaction, and nutrient prediction.

\section{Methodology}
\label{sec:method}

NutriVision estimates the nutritional composition of a dish by combining three complementary sources of evidence. The \textit{RGB image} captures appearance and food identity; the \textit{estimated depth map} provides geometric cues related to portion size and spatial structure; and \textit{the ingredient list} supplies explicit information about food composition. As shown in Figure~\ref{fig:pipeline}, the model first corrects the monocular depth representation using an SSRA. RGB and adapted-depth inputs are then processed by separate ConvNeXt-base encoders and refined through modality-specific ISM blocks. The resulting feature pyramids are fused in the frequency domain under ingredient guidance, aggregated across scales, and passed to an IA-MPH. The complete model is trained using a normalized nutrition-regression objective together with an auxiliary RGB-depth alignment loss.

\subsection{Problem Formulation and Notation}
Given an RGB image $\mathbf{I}\in\mathbb{R}^{3\times224\times224}$, an
ingredient string $s$, a pretrained monocular depth estimator produces a depth map,
$\mathbf{D}^{\text{mono}}= g_{\mathrm{depth}}(\mathbf{I})\in\mathbb{R}^{1\times224\times224}$ where $= g_{\mathrm{depth}}$ denotes DepthAnything-V3. The ingredient string is encoded offline using the CLIP ViT-B/32 text encoder, $\mathbf{e}
= g_{\mathrm{text}}(s)
\in\mathbb{R}^{512}.$. Following IGSMNet~\cite{foods14213697}, the experiments use ground-truth ingredient strings supplied with Nutrition5k. The ingredient information is therefore treated as an input available at inference time rather than as a target predicted by NutriVision. The model predicts a 5-dimensional nutrition vector
\begin{equation}
\hat{\mathbf{y}} = (\hat{y}_{\text{cal}},\hat{y}_{\text{mass}},
        \hat{y}_{\text{fat}},\hat{y}_{\text{carb}},
        \hat{y}_{\text{prot}})\in\mathbb{R}^{5}.
\label{eq:output}
\end{equation}
\noindent that contains calories, total mass, fat, carbohydrates, and protein. The complete mapping can be written as $F_{\Theta}
\left(
\mathbf{I},
\mathbf{D}^{\mathrm{mono}},
\mathbf{e}
\right)$ where $(\Theta)$ denotes the trainable parameters of the depth adapter, dual visual encoders, fusion modules, and prediction head. 
Algorithm~\ref{alg:forward} summarizes the full forward pass.

\begin{algorithm}[t]
\caption{NutriVision Forward Pass}
\label{alg:forward}
\begin{algorithmic}[1]
\Require RGB image $I$, ingredient string $s$, monocular depth $D_{\text{mono}}$
\Ensure Nutrition vector $\hat{y} \in \mathbb{R}^5$

\State $e \gets \text{CLIP-Text}(s)$ \Comment{offline embedding}
\State $D \gets \text{SSRA}(D_{\text{mono}})$ \Comment{Eqs.~2--4}

\For{$i \in \{0,1,2,3\}$} \Comment{per-scale pyramid}
    \State $r_i \gets \text{ConvNeXt-RGB}_i(I)$
    \State $d_i \gets \text{ConvNeXt-Depth}_i(D)$
    \State $r_i \gets \text{ISM}(r_i)$; \quad $d_i \gets \text{ISM}(d_i)$ \Comment{Eq.~5}
    \State $O_i \gets \text{IC-FAFM}(r_i, d_i, e)$ \Comment{Eqs.~6--14}
\EndFor

\State $f \gets \text{MultiScaleAggregate}(\{O_i\}_{i=0}^{3})$ \Comment{Eqs.~12--14}
\State $c_r, c_d \gets \text{GlobalPool}(r_3), \text{GlobalPool}(d_3)$
\State $x \gets [f\, ; c_r\, ; c_d]$ \Comment{Eq.~16}
\State $x \gets x \odot \sigma(\text{Linear}(e))$ \Comment{Eq.~17, ingredient gate}

\If{training}
    \State $m \gets \mathbf{1}[|x| \geq \text{quantile}_{0.25}(|x|)]$
    \State $x \gets x \odot m$ \Comment{Eq.~18, channel mask, train-only}
\EndIf

\State $\hat{y} \gets \text{IA-MPH-Head}(x, e)$ \Comment{cross-attn + MLP, Sec.~III-F}
\State \Return $\hat{y}$
\end{algorithmic}
\end{algorithm}

\subsection{Scale-Shift Residual Adapter (SSRA)}
Monocular depth is defined up to an unknown affine transformation. Following
OmniFood8k~\cite{yu2026omnifood8k}, we apply a global affine correction
followed by a local CNN residual:
\begin{align}
\mathbf{D}^{\text{glo}} &= \alpha\,\mathbf{D}^{\text{mono}} + \beta,
   \label{eq:ssra1}\\
\mathbf{D}^{\text{res}} &= f_\theta(\mathbf{D}^{\text{glo}}),
   \label{eq:ssra2}\\
\mathbf{D} &= \mathbf{D}^{\text{glo}} + \mathbf{D}^{\text{res}}.
   \label{eq:ssra3}
\end{align}
Here $\alpha,\beta\in\mathbb{R}$ are learnable scalars (init.\ $1,0$) and
$f_\theta$ is a four-layer CNN
($1\!\to\!64\!\to\!64\!\to\!32\!\to\!1$, $3{\times}3$ kernels, BN, ReLU).
SSRA contributes $\sim$100\,K parameters.

\subsection{Dual ConvNeXt-Base Backbone}
Both modalities are encoded by ConvNeXt-Base pretrained on
ImageNet-22k and fine-tuned on ImageNet-1k~\cite{liu2022convnet}; a
$\mathrm{Conv2d}(1{\to}3)$ adapter (weights $1/3$) precedes the depth
branch. Each branch yields a four-stage pyramid with strides
$\{4,8,16,32\}$ and channels $\{128,256,512,1024\}$, projected to a
common 256 channels by $1{\times}1$ convolution. The first two stages
are frozen, leaving $\sim$44\,M trainable parameters per branch.
Global CLS embeddings $\mathbf{c}_r,\mathbf{c}_d\in\mathbb{R}^{512}$
come from $\mathrm{AdaptiveAvgPool2d}(1){\to}\mathrm{Linear}(1024,512){\to}\mathrm{ReLU}$.
We chose ConvNeXt-Base over IGSMNet's Swin-Tiny for its larger
capacity and stronger ImageNet-22k pretraining.

\subsection{Internal Semantic Modelling (ISM)}
We apply IGSMNet's ISM block~\cite{foods14213697} symmetrically to both
modalities, independently per scale. Each block combines (i) a Dynamic
Position Encoding (DPE) that maps relative window offsets to an
attention bias, $b_{ij}=\mathrm{MLP}_{\text{DPE}}(\Delta x_{ij},\Delta y_{ij})$,
and (ii) Fine-Grained Modelling (FGM), window-based multi-head
self-attention with that bias:
\begin{equation}
\mathrm{FGM}(\mathbf{X}) = \mathrm{softmax}\!\left(
\tfrac{\mathbf{QK}^{\!\top}}{\sqrt{d}}+\mathbf{B}\right)\mathbf{V},
\label{eq:fgm}
\end{equation}
on $7{\times}7$ windows, followed by a pre-norm transformer MLP with
residuals. We use two ISM blocks per scale per modality (16 total).

\subsection{Ingredient-Conditioned Frequency-Aligned Fusion (IC-FAFM)}
\label{sec:icfafm}
IC-FAFM extends OmniFood8K's Frequency-Aligned Fusion Module
(FAFM)~\cite{yu2026omnifood8k} with multi-scale operation and
ingredient-conditioned cross-attention; we retain the FAFM acronym to
acknowledge this lineage. The module is applied at each of the four
ConvNeXt scales $i{\in}\{0,1,2,3\}$. The individual module-level
deltas (A7, A8, A11 in Tab.~\ref{tab:module_ablation}) sit at the
per-run noise floor, so we treat IC-FAFM as a small, diffuse
contributor.

\begin{figure}[!t]
\centering
\includegraphics[width=\columnwidth]{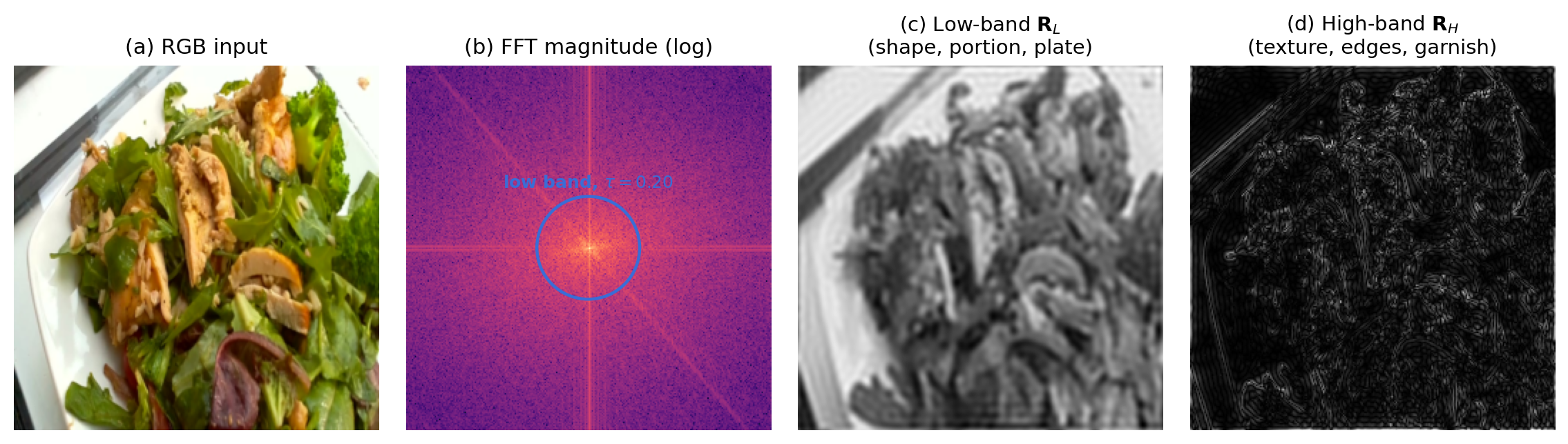}
\caption{What the FFT split inside IC-FAFM separates. (a)~the input dish.
(b)~its log-magnitude 2-D FFT, with the low-pass mask of radius
$\tau\,r_{\max}=0.20\,r_{\max}$ overlaid. (c)~inverse FFT of the low band,
which preserves plate boundary, portion footprint, and overall layout.
(d)~inverse FFT of the high band, which preserves edges, textures, and
fine-grained ingredient appearance.}
\label{fig:fft}
\end{figure}

\paragraph{Step~1: Frequency-Aligned Fusion}
Let $\mathbf{r}_i,\mathbf{d}_i\in\mathbb{R}^{B\times 256\times H_i\times W_i}$
denote the post-ISM RGB and depth features. We compute the 2-D FFT of each:
\begin{equation}
\mathbf{R}_f = \mathcal{F}(\mathbf{r}_i),\quad
\mathbf{D}_f = \mathcal{F}(\mathbf{d}_i).
\label{eq:fft}
\end{equation}
A circular low-pass mask $\mathbf{M}_L$ of radius
$\tau\!\cdot\!r_{\max}$ ($\tau{=}0.20$) decomposes each feature into a
low-frequency and a high-frequency component:
\begin{align}
\mathbf{R}_L &= \mathcal{F}^{-1}(\mathbf{R}_f\!\odot\!\mathbf{M}_L),
&\mathbf{R}_H &= \mathcal{F}^{-1}(\mathbf{R}_f\!\odot\!(\mathbf{1}-\mathbf{M}_L)),\\
\mathbf{D}_L &= \mathcal{F}^{-1}(\mathbf{D}_f\!\odot\!\mathbf{M}_L),
&\mathbf{D}_H &= \mathcal{F}^{-1}(\mathbf{D}_f\!\odot\!(\mathbf{1}-\mathbf{M}_L)).
\end{align}
The two modalities are added per band and merged by a $3{\times}3$ convolution:
\begin{equation}
\mathbf{O}_i = \mathrm{Conv}_{3\times 3}\!\left([\,
   \mathbf{R}_L+\mathbf{D}_L\,;\,\mathbf{R}_H+\mathbf{D}_H\,]\right).
\label{eq:fafm}
\end{equation}
Figure~\ref{fig:fft} visualises what this split actually buys: at $\tau{=}0.20$ the
low band keeps the food's footprint and plate geometry, while the high band
isolates texture and edge information---the cues most useful for distinguishing
sauces, garnishes, and surface fat.

\paragraph{Hyperparameter $\tau$}
$\tau{=}0.20$ is inherited from OmniFood8k's FAFM. We did not run a
sweep; since A8 (no FFT split) is at the noise floor, any reasonable
$\tau$ should move the mean by less. A sweep and scale-adaptive
$\tau_i$ are future work.

\begin{figure}[!t]
\centering
\includegraphics[width=\columnwidth]{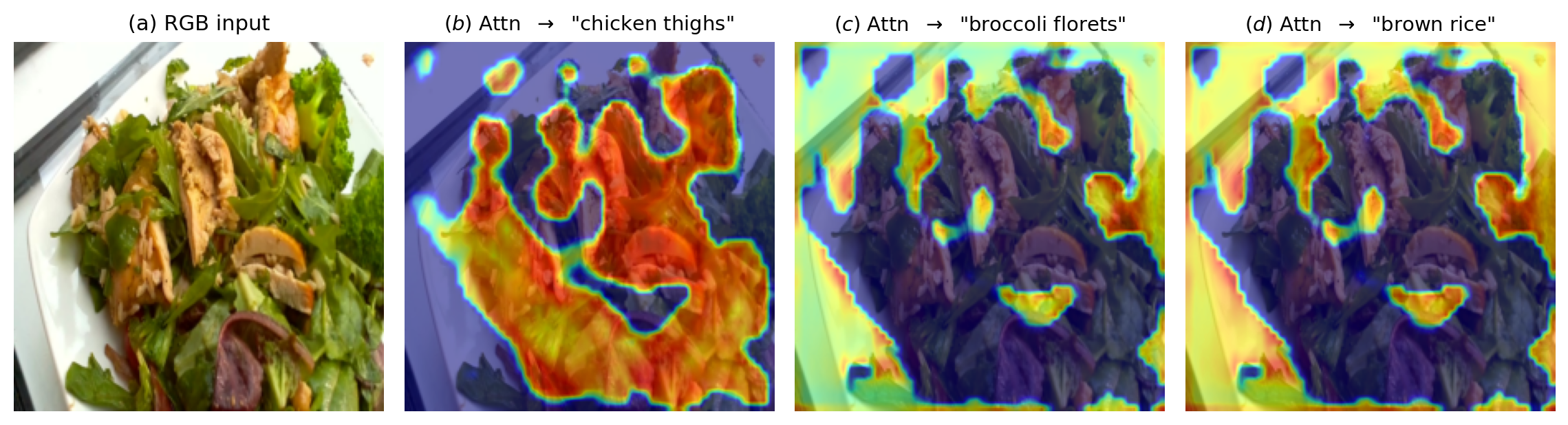}
\caption{Ingredient cross-attention inside IC-FAFM (Eq.~\ref{eq:xattn}) on the
same dish, queried with three different ingredient strings. The hot regions
shift meaningfully: ``chicken thighs'' attends to the seared chicken on the
right, ``broccoli florets'' migrates upward toward the green florets, and
``brown rice'' redistributes again. The attention is a soft prior, not a
segmentation; this is \emph{qualitative evidence of ingredient--visual
grounding}, not causal evidence of a headline accuracy gain---the
corresponding retraining ablation (A7, $\Delta\!=\!+0.18$\,PMAE) is at
the per-run noise floor.}
\label{fig:attn}
\end{figure}

\paragraph{Step~2: Ingredient Cross-Attention}
The ingredient embedding is projected to the feature dimension,
$\mathbf{q}=\mathrm{Linear}(\mathbf{e})\in\mathbb{R}^{B\times 256}$, and
used as a single query against the spatial tokens of $\mathbf{O}_i$
via standard scaled dot-product attention~\cite{vaswani2017attention}:
\begin{equation}
\mathbf{a}_i = \mathrm{MHA}\!\Big(\mathbf{q}^{\!\top},\;\Phi(\mathbf{O}_i),\;\Phi(\mathbf{O}_i)\Big),
\label{eq:xattn}
\end{equation}
with $\Phi$ flattening to spatial tokens and four attention heads. The
result is broadcast back to the spatial grid:
\begin{equation}
\tilde{\mathbf{O}}_i = \mathbf{O}_i + \mathrm{expand}(\mathbf{a}_i;H_i,W_i).
\label{eq:guided}
\end{equation}
Figure~\ref{fig:attn} shows the attention shifting with the queried
ingredient---qualitative evidence of grounding, not causal evidence
of the headline gain (A7 is $+0.18$\,PMAE, at the noise floor),
consistent with the small-and-diffuse framing of IC-FAFM
(Sec.~\ref{sec:module_ablation}).

\paragraph{Step~3: Multi-Scale Aggregation}
Operating over four scales follows the now-standard feature-pyramid
recipe~\cite{lin2017feature}: per-scale pooled descriptors are sigmoid-gated and
concatenated:
\begin{align}
\mathbf{p}_i &= \mathrm{ReLU}(\mathrm{Linear}(\mathrm{AvgPool}(\tilde{\mathbf{O}}_i))),\\
\mathbf{g}_i &= \sigma(s_i)\,\mathbf{p}_i,\\
\mathbf{f}   &= \mathrm{MLP}\!\big([\mathbf{g}_0;\mathbf{g}_1;\mathbf{g}_2;\mathbf{g}_3]\big)
                \in\mathbb{R}^{B\times 256},
\label{eq:msagg}
\end{align}
where $s_i\!\in\!\mathbb{R}$ are learnable per-scale gates. The MLP has
shape $1024{\to}512{\to}256$.

\paragraph{Step~4: Inter-Modal Alignment Loss}
For each scale we project the pre-fusion features through small heads
$\psi_r,\psi_d:\mathbb{R}^{256}\!\to\!\mathbb{R}^{128}$ and apply an
info-NCE contrastive loss across the batch:
\begin{equation}
\mathcal{L}_{\text{align}}^{(i)} = -\frac{1}{B}\sum_{b=1}^{B}\log
\frac{\exp(\langle\psi_r(\mathbf{r}_i^{(b)}),\psi_d(\mathbf{d}_i^{(b)})\rangle/\tau_c)}
{\sum_{b'=1}^{B}\exp(\langle\psi_r(\mathbf{r}_i^{(b)}),\psi_d(\mathbf{d}_i^{(b')})\rangle/\tau_c)}.
\label{eq:nce}
\end{equation}
The four scale losses are averaged into~$\mathcal{L}_{\text{align}}$.

\subsection{Ingredient-Aware Mask-based Prediction Head (IA-MPH)}
\label{sec:iamph}
\emph{IA-MPH is our second novel contribution.} The fused vector
$\mathbf{f}\!\in\!\mathbb{R}^{256}$ is concatenated with the CLS embeddings
of both modalities,
\begin{equation}
\mathbf{x} = [\mathbf{f};\mathbf{c}_r;\mathbf{c}_d]\in\mathbb{R}^{1280}.
\end{equation}

\paragraph{Ingredient-conditioned gating}
\begin{equation}
\mathbf{x} \leftarrow \mathbf{x}\;\odot\;\sigma(\mathrm{Linear}_{512\to 1280}(\mathbf{e})).
\label{eq:gate}
\end{equation}

\paragraph{Dynamic channel mask (training only)} Following the masked-channel
philosophy of~\cite{yu2026omnifood8k}, we retain the top~75\% of channels by
magnitude:
\begin{equation}
\mathbf{m} = \mathbf{1}\!\left[|\mathbf{x}|\ge\mathrm{quantile}_{0.25}(|\mathbf{x}|)\right],
\quad \mathbf{x}\leftarrow\mathbf{x}\odot\mathbf{m}.
\label{eq:mask}
\end{equation}
This regulariser encourages the network to commit informative channels to
each food type.

\paragraph{Ingredient cross-attention and regression}
A second cross-attention block uses $\mathbf{x}$ as query against the
ingredient embedding, followed by a gated fusion and a three-layer MLP head
$1280{\to}512{\to}256{\to}5$ with dropouts $0.1$ and~$0.05$. The output is the
5-D nutrition vector~\eqref{eq:output}.

\subsection{Loss Function}
We minimise
\begin{equation}
\mathcal{L} \;=\; \mathcal{L}_{\text{nutri}} + \lambda_a\,\mathcal{L}_{\text{align}},
\quad \lambda_a = 0.15,
\label{eq:total_loss}
\end{equation}
where $\mathcal{L}_{\text{align}}$ is the average of Eq.~\eqref{eq:nce} over
the four scales, and $\mathcal{L}_{\text{nutri}}$ is the
\emph{normalised}~$\ell_1$ loss of~\cite{foods14213697}:
\begin{equation}
\mathcal{L}_{\text{nutri}}=\sum_{t=1}^{5}
\frac{\sum_{b=1}^{B}\big|\hat{y}_{t,b}-y_{t,b}\big|}
     {\sum_{b=1}^{B} y_{t,b}}.
\label{eq:nutri_loss}
\end{equation}
This loss directly optimises the PMAE metric and is scale-invariant across
the five tasks.

\subsection{Parameter Budget}
The full model has $\sim$196\,M parameters of which $\sim$105\,M are trainable
(Table~\ref{tab:params}).

\begin{table}[!t]
\caption{Per-component parameter budget.\label{tab:params}}
\centering
\scriptsize
\setlength{\tabcolsep}{3pt}
\begin{tabular}{lrrl}
\toprule
Component & Params & Trainable & Notes\\
\midrule
ConvNeXt-Base RGB        & 88.6\,M & 44\,M  & Stages 2--3 unfrozen\\
ConvNeXt-Base Depth      & 88.6\,M & 44\,M  & Stages 2--3 unfrozen\\
Depth adapter            & 6       & 6      & $1\!\to\!3$ Conv$2\times2$\\
Scale projections (8)    & 790\,K  & 790\,K & $1{\times}1$ to 256-ch\\
CLS projections (2)      & 1.1\,M  & 1.1\,M & Linear $1024{\to}512$\\
SSRA                     & 100\,K  & 100\,K & $\alpha,\beta$ + CNN refiner\\
ISM RGB ($2{\times}4$)   & 3.2\,M  & 3.2\,M & Window attn + MLP\\
ISM Depth ($2{\times}4$) & 3.2\,M  & 3.2\,M & Window attn + MLP\\
IC-FAFM (4 scales)       & 8.0\,M  & 8.0\,M & FAFM + cross-attn + align.\\
IA-MPH                   & 2.0\,M  & 2.0\,M & Gate + cross-attn + reg.\\
Alignment projectors     & 530\,K  & 530\,K & $4{\times}(256{\to}128){\times}2$\\
\midrule
\textbf{Total}           & \textbf{$\sim$196\,M} & \textbf{$\sim$105\,M} & \\
\bottomrule
\end{tabular}
\end{table}

\begin{table}[!t]
\caption{Compute budget for NutriVision. Forward-pass FLOPs are
\emph{estimated} from per-block published numbers (ConvNeXt-Base
$\sim$15.4\,GFLOPs at $224^2$~\cite{liu2022convnet}); training and
inference times are \emph{measured} on a single RTX A6000 (48\,GB).
End-to-end inference includes the DepthAnything-V3 forward and the
CLIP text encoding for the ingredient string.\label{tab:compute}}
\centering
\scriptsize
\setlength{\tabcolsep}{3pt}
\begin{tabular}{@{}l r p{1.45in}@{}}
\toprule
Quantity & Value & Notes\\
\midrule
Total parameters       & $\sim$196\,M       & measured (Tab.~\ref{tab:params})\\
Trainable parameters   & $\sim$105\,M       & stages 2--3 of both backbones unfrozen\\
Forward FLOPs (est.)   & $\sim$35\,GFLOPs   & $2{\times}15.4$ (backbones) $+$ ISM/IC-FAFM/IA-MPH/SSRA overhead\\
Train time / epoch     & $\sim$925\,s       & batch 8 on RTX A6000, 1923 batches/epoch\\
Total train time       & $\sim$38.5\,h      & 150 epochs\\
End-to-end inference   & $\sim$3\,s / image & RGB$\to$depth$\to$CLIP$\to$NutriVision\\
Model forward only     & $<$0.5\,s / image  & batch 1, RTX A6000 (depth+CLIP precomputed)\\
\bottomrule
\end{tabular}
\end{table}

\paragraph{Note on baseline scale}
Tab.~\ref{tab:sota} is \emph{not} parameter-matched. Kwan
NuNet~\cite{kwan2025nutrition} (dual SwinT-Base, $\sim$200--260\,M) and
OmniFood8k~\cite{yu2026omnifood8k} (dual ConvNeXt-Base) are broadly
comparable to NutriVision ($\sim$196\,M); IGSMNet~\cite{foods14213697}
uses a smaller Swin-family backbone, and the earlier baselines
(Inception-V2, Swin-Nutrition, DPF-Nutrition) are smaller still.
Two consequences follow. First, the $\sim$5--7\,pp mean-PMAE gap to the
smaller baselines should be read as a \emph{combined effect of
architecture, ingredient input, and backbone scale}, not as pure
architectural superiority. Second, the input-matched comparison against
IGSMNet (re-run, $0.89$\,pp) is also \emph{not} parameter-matched: a
size-controlled IGSMNet (e.g.\ with a Swin-Base or ConvNeXt-Base
backbone) is left to future work
(Sec.~\ref{sec:discussion}).

\section{Experiments}
\label{sec:experiments}

\subsection{Dataset and Protocol}
We evaluate on \textbf{Nutrition5k}~\cite{thames2021nutrition5k}, which
contains 5,006 unique dishes captured from four overhead cameras
(A/B/C/D) together with per-dish ground-truth nutrition. Each dish
provides $\sim$4 RGB views and one calibrated depth view. We follow the
\emph{5:1 dish-level split} of~\cite{foods14213697}, yielding approximately
15{,}000 training and 3{,}000 testing images. The split is performed at the
dish level so that camera views of the same dish never leak between train
and test.

\paragraph{Best-on-test selection (protocol disclosure)}
We use the test set both for early stopping and for reporting (no
separate held-out validation split); every entry in
Tabs.~\ref{tab:sota}--\ref{tab:module_ablation} is a
\emph{best-mean-PMAE-on-test} value. This matches IGSMNet's and
Kwan~\textit{et~al.}'s protocol so within-table comparisons are
like-for-like, but the headline $13.60\pm0.10\%$ is optimistically
biased relative to true held-out generalisation; a proper held-out
split or $k$-fold CV is left to Sec.~\ref{sec:discussion}.

Monocular depth maps are precomputed once with
DepthAnything-V3~\cite{lin2025depth}; CLIP ingredient embeddings
are precomputed with the ViT-B/32 text encoder. We additionally precompute
food-vs-background masks with SAM~\cite{kirillov2023segment} using the
``food on a plate'' text prompt; these masks are retained as a preprocessing
artifact and explored in Sec.~\ref{sec:discussion} but are \emph{not}
consumed by the reported model, since the empirical contribution of
foreground masking was negative in our pilot experiments. No data
augmentation is used beyond a synchronised horizontal flip ($p{=}0.5$)
applied jointly to RGB ($\mathbf{I}\in\mathbb{R}^{3\times 224\times 224}$)
and depth.

\paragraph{Test-time input asymmetry}
NutriVision consumes RGB$+$estimated depth$+$GT ingredient string;
the ingredient input is the same privileged signal IGSMNet uses.
The input-matched comparison in Tab.~\ref{tab:sota} is therefore
with IGSMNet, and lower-mean claims against ingredient-free
baselines (OmniFood8k, DPF-Nutrition, Google-Nutrition-rgbd,
Kwan~\textit{et~al.}) should be read as ``best-in-class given the
privileged input.'' Removing CLIP inflates NutriVision to $27.44\%$
(A2, $+13.92$).

\subsection{Evaluation Metric}
We report Percentage Mean Absolute Error (PMAE) per task and its mean
over the five tasks:
\begin{equation}
\mathrm{PMAE}_t = \frac{\tfrac{1}{N}\sum_{n}|\hat{y}_{t,n}-y_{t,n}|}
                       {\tfrac{1}{N}\sum_{n}y_{t,n}}\times 100\%,
\label{eq:pmae}
\end{equation}
\begin{equation}
\mathrm{Mean\,PMAE} = \frac{1}{5}\sum_{t=1}^{5}\mathrm{PMAE}_t.
\label{eq:meanpmae}
\end{equation}

\subsection{Implementation Details}
\begin{figure}[!t]
\centering
\includegraphics[width=\linewidth]{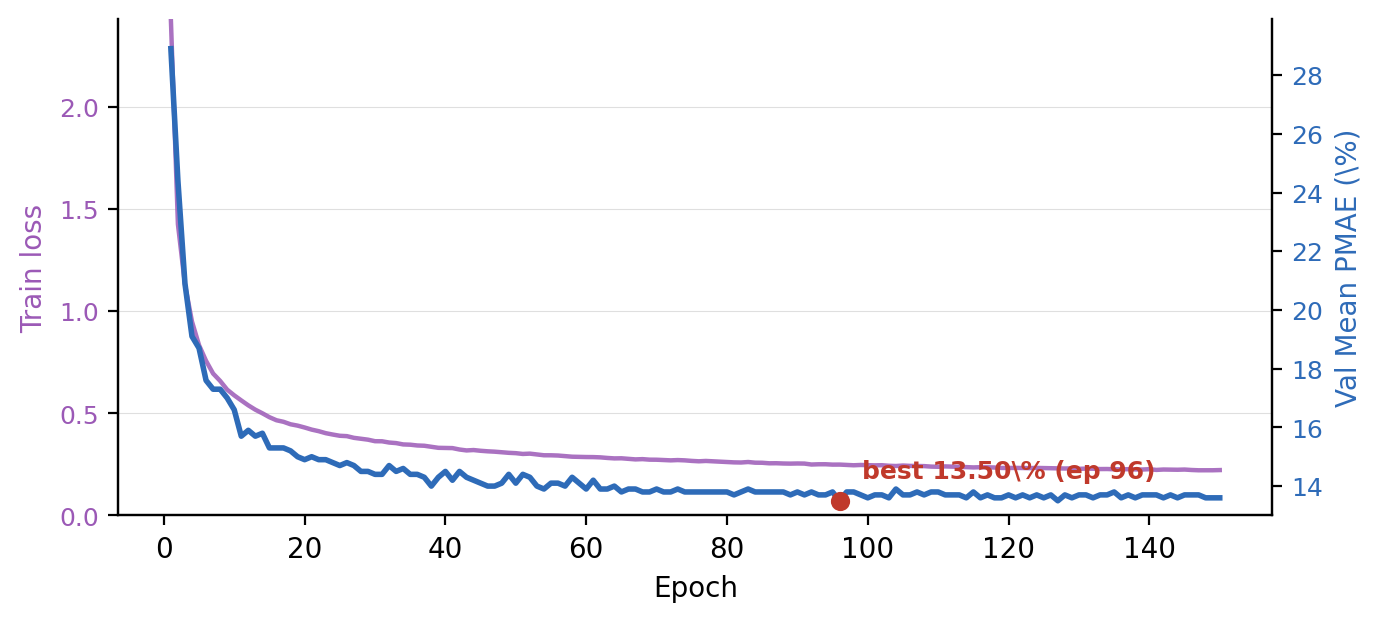}
\caption{Training trajectory on Nutrition5k (representative seed).
Train loss (purple, left) decays smoothly; test-set mean PMAE (blue,
right) plateaus near $13.5\%$; we report the best-on-test checkpoint
(\S\ref{sec:experiments}). The 3-seed mean best-mean-PMAE is
$13.60\pm0.10\%$.}
\label{fig:training}
\end{figure}

We use Adam with a two-tier learning rate of $1{\rm e}{-5}$ for the unfrozen
ConvNeXt stages and $5{\rm e}{-5}$ for the rest; exponential schedule with
$\gamma=0.99$ per epoch; weight decay~$1{\rm e}{-5}$; gradient clipping at
$\|\cdot\|_2\!=\!1$. Batch size is~8, training runs for 150 epochs, on a
single NVIDIA RTX~A6000 (48\,GB). End-to-end training takes
$\sim$925\,s/epoch, i.e.\ $\sim$38.5\,h total. Hyperparameters are summarised
in Table~\ref{tab:hparams}. The training trajectory
(Fig.~\ref{fig:training}) is shown for a representative seed and
converges to a best mean PMAE of $13.50\%$ (3-seed mean
$13.60\pm0.10\%$) on the Nutrition5k test split, after which the model
plateaus.
Selection uses the same test split throughout
(\S\ref{sec:experiments}); no held-out validation split is carved
out.

\begin{table}[!t]
\caption{Training hyperparameters.\label{tab:hparams}}
\centering
\footnotesize
\begin{tabular}{ll}
\toprule
Parameter & Value\\
\midrule
Optimizer                & Adam\\
Base learning rate       & $5{\rm e}{-5}$\\
Backbone learning rate   & $1{\rm e}{-5}$\\
LR schedule              & Exp.\ decay, $\gamma=0.99$/epoch\\
Weight decay             & $1{\rm e}{-5}$\\
Gradient clipping        & $\|\cdot\|_2 = 1.0$\\
Batch size               & 8\\
Epochs                   & 150\\
Input resolution         & $224\!\times\!224$\\
Augmentation             & Sync.\ horizontal flip ($p{=}0.5$)\\
CLIP model               & ViT-B/32\\
FAFM threshold $\tau$    & 0.20\\
ISM blocks per level     & 2\\
ISM window size          & $7\!\times\!7$\\
Alignment weight $\lambda_a$ & 0.15\\
\bottomrule
\end{tabular}
\end{table}

\section{Results}
\label{sec:results}

\subsection{Comparison with State-of-the-Art}
Table~\ref{tab:sota} compares NutriVision with six published
baselines and our reimplementation of IGSMNet (full training code
was not publicly available, so we re-implemented from the paper's
architectural description and trained under identical conditions to
NutriVision). NutriVision attains a 3-seed mean PMAE of
$\mathbf{13.60\pm0.10\%}$, the lowest in the table.

\paragraph{Statistical significance}
The headline NutriVision configuration and the dominant module ablation
(A10, IA-MPH$\rightarrow$plain MLP) were re-trained under identical
conditions across three seeds $\{1,2,3\}$ affecting model
initialisation, dropout, and stochastic data ordering (the dish-level
5{:}1 split is fixed). NutriVision reaches
$13.60\pm0.10\%$ (per-seed: $13.50,13.60,13.70$) and A10 reaches
$15.10\pm0.26\%$ ($14.90,15.00,15.40$). A one-sample $t$-test of the
NutriVision seeds against the IGSMNet reimplementation ($14.49\%$)
rejects the null at $t\!=\!-15.4$, $p\!<\!0.001$; against IGSMNet's
published $15.0\%$ at $t\!=\!-24.3$, $p\!<\!0.001$. A paired $t$-test
of A10 against NutriVision yields
$\Delta\!=\!+1.50\pm0.17$\,pp, $p\!<\!0.001$. Rows A7--A9, A11, A12
remain single-seed and their $|\Delta|\!\le\!0.20$ deltas continue to
sit inside the $\sim$$0.15$\,pp per-run noise floor; scaling the
paired protocol to those rows is left as future work.

\paragraph{What the comparison fairly establishes}
The input-matched comparison is against IGSMNet: $+0.89\pm0.10$\,pp vs
our re-run ($14.49\!\to\!13.60\pm0.10$, $p\!<\!0.001$),
$+1.40\pm0.10$\,pp vs IGSMNet's reported numbers
($15.0\!\to\!13.60\pm0.10$, $p\!<\!0.001$). This is the principal
claim. Comparisons to ingredient-free baselines (OmniFood8k,
DPF-Nutrition, Google-Nutrition-rgbd, Kwan~\textit{et~al.}) are not
input-matched---A2 shows that removing ingredients alone pushes
NutriVision to $27.44\%$, worse than every ingredient-free baseline---so
cross-input gaps characterise an ingredient-aware operating point, not
architectural superiority.

\begin{table*}[!t]
\caption{Per-nutrient and mean PMAE (\%) on the Nutrition5k 5{:}1
dish-level split. Lower is better. Best in bold. $^{\dagger}$The
``IGSMNet (our reimpl.)'' row is our re-implementation of the IGSMNet
architecture from~\cite{foods14213697} (the full training code was
not publicly available at the time of writing); we follow the paper's
architectural description and train under identical conditions to
NutriVision. $^{\ddagger}$Per-nutrient columns for the NutriVision row
are from a representative seed=1 run (best-mean-PMAE $13.50\%$); the
Mean column reports the 3-seed mean$\pm$std, and the $\Delta$ column
uses the 3-seed mean. All baselines are as published.
\label{tab:sota}}
\centering
\footnotesize
\begin{tabular}{lllccccccc}
\toprule
Method & Venue & Input (test) & Cal & Mass & Fat & Carb & Prot & Mean & $\Delta$ vs.\ ours\\
\midrule
Google-Nutrition-rgbd~\cite{thames2021nutrition5k} & CVPR'21 & RGB-D (real) & 18.8 & 18.9 & 18.1 & 23.8 & 20.9 & 20.1 & $+6.50$\\
Swin-Nutrition~\cite{foods11213429}                  & Foods'22 & RGB         & 15.3 & 12.5 & 22.1 & 20.8 & 15.4 & 17.2 & $+3.60$\\
DPF-Nutrition~\cite{foods12234293}                    & Foods'23 & RGB (est.\ depth) & 14.7 & 10.6 & 22.6 & 20.7 & 20.2 & 17.8 & $+4.20$\\
OmniFood8k~\cite{yu2026omnifood8k}                   & CVPR'26  & RGB         & 14.1 & 10.2 & 21.0 & 18.9 & 18.4 & 16.5 & $+2.90$\\
Kwan~\textit{et~al.}~\cite{kwan2025nutrition} (NuNet)                       & TMM'25   & RGB-D (real)& 12.80 & 8.72 & 19.67 & 18.66 & 18.42 & 15.65 & $+2.05$\\
IGSMNet~\cite{foods14213697}                            & Foods'25 & RGB-D (real)+Ingr.   & 12.2 & 9.4 & 19.1 & 18.3 & 16.0 & 15.0 & $+1.40$\\
IGSMNet (our reimpl.)$^{\dagger}$\,\cite{foods14213697}     & --       & RGB-D (real)+Ingr.   & 13.6 & 11.4 & 17.6 & 15.5 & 14.4 & 14.49 & $+0.89$\\
\midrule
\textbf{NutriVision (ours)}$^{\ddagger}$              & --       & RGB+Ingr.+Depth(est.) & \textbf{12.65} & \textbf{10.37} & \textbf{16.45} & \textbf{14.76} & \textbf{13.35} & $\mathbf{13.60\!\pm\!0.10}$ & --\\
\bottomrule
\end{tabular}
\end{table*}

\begin{figure}[!t]
\centering
\includegraphics[width=\columnwidth]{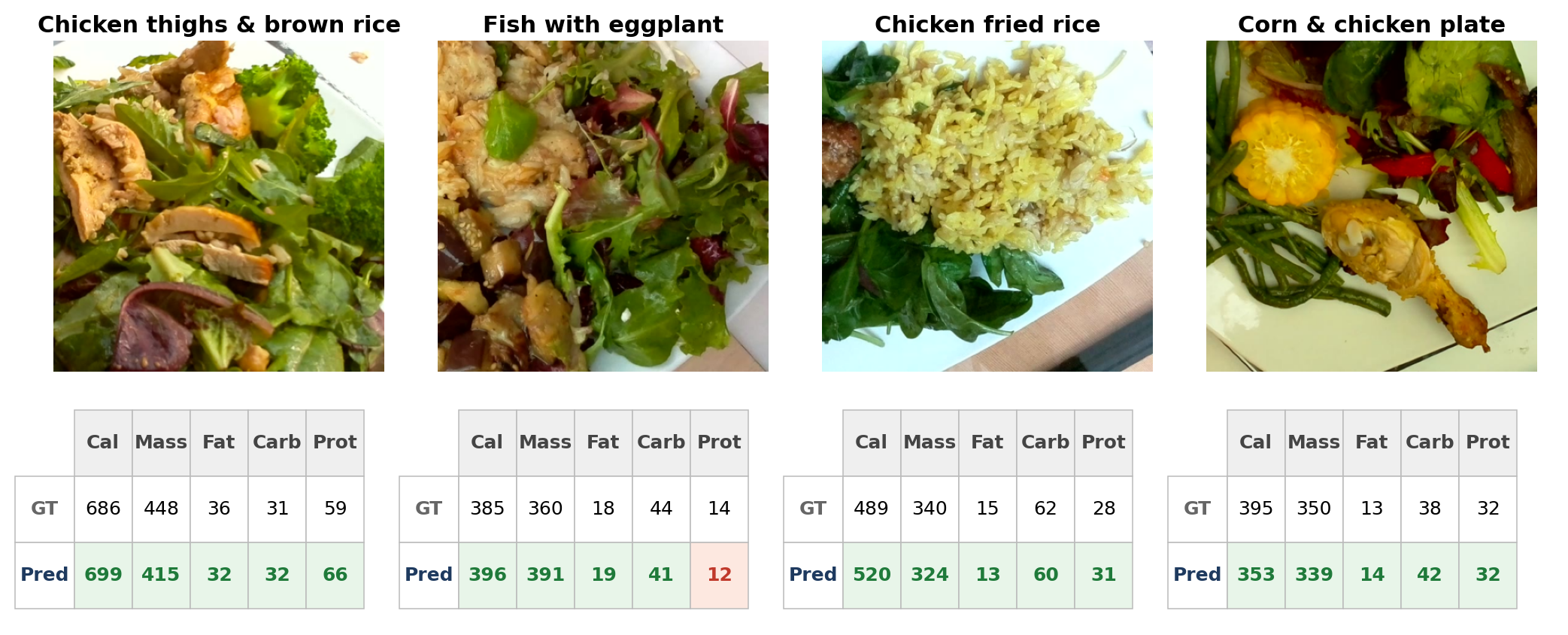}
\caption{Four representative Nutrition5k test predictions, spanning low-,
mid-, and high-calorie dishes. For each plate we show the input photograph,
the comma-separated ingredient list used at inference, and a compact
ground-truth-vs-prediction table for the five nutrients. Green cells
indicate the prediction is within $\pm$13.5\% of GT (one PMAE band);
orange cells mark larger misses. Errors are largest on dishes with
heavily-stacked or occluded ingredients.}
\label{fig:qualitative}
\end{figure}

Compared with our IGSMNet reimplementation, NutriVision improves on every
per-nutrient metric; largest gains: fat ($-1.15$, $17.6\!\to\!16.45$)
and protein ($-1.05$, $14.4\!\to\!13.35$). Against \emph{published}
numbers the per-nutrient story is mixed: NutriVision loses on mass to
Kwan ($8.72$), IGSMNet ($9.4$), and OmniFood8k ($10.2$)---all below
our $10.37$. The pattern is physically sensible (real sensor depth
and larger ingredient-aware backbones retain a mass advantage over
our DepthAnything-V3 input). NutriVision wins on the mean against every
published baseline and on four of five nutrients individually. Figure~\ref{fig:qualitative}
shows six representative test-set predictions; the model produces
within-$\pm$13.5\,\% estimates on the majority of nutrients per dish, with
the largest errors concentrated on visually-ambiguous mixed plates.

\paragraph{What the ablations actually attribute the gains to}
The per-nutrient gains over IGSMNet cannot be plausibly credited to
SSRA or FFT: A6 moves mean by $0.00$, A3 moves mass by $+0.05$ only,
and A8 moves fat by $+0.35$---all an order of magnitude smaller than
the $\sim$1\,pp fat/mass gains. The supported drivers are (i)~the
IA-MPH head (A10, $+1.50\pm0.17$\,pp, $p\!<\!0.001$) and (ii)~the
backbone swap from Swin-family to ConvNeXt-Base (inferred from
IGSMNet$\to$NutriVision moves; a formal backbone ablation is future
work). SSRA, FFT split, and the alignment loss should therefore be
read as scaffolding that may contribute jointly but whose individual
deltas are at or just above the per-run noise floor.

\subsection{Comparison with Real-Depth Methods}
NutriVision ($13.60\pm0.10\%$) is numerically lower than
Kwan~\textit{et~al.}'s NuNet~\cite{kwan2025nutrition} (15.65\%), the
strongest published method using the real depth captures of
Nutrition5k. However, this is \emph{not} an apples-to-apples depth
comparison: NuNet operates on RGB+depth with no ingredient signal,
while NutriVision combines RGB, estimated depth, and CLIP-encoded
ingredient text. Our input ablation (Tab.~\ref{tab:input_ablation},
A3) shows that the depth pathway contributes only $+0.13$\,PMAE when
zeroed at inference (a test-time robustness measurement, not a
from-scratch retraining without depth---see the caveat in
Sec.~\ref{sec:ablation}), and the module ablation
(Tab.~\ref{tab:module_ablation}) attributes the dominant gains to the
ingredient-aware prediction head (A10, $+1.50\pm0.17$, $p\!<\!0.001$)
and to the metric-aligned normalised-$\ell_1$ loss adopted from
IGSMNet~\cite{foods14213697} (A12, $+1.20$, single seed; a
confirmation of an inherited choice rather than a novel contribution).
The fair reading: \emph{NutriVision with estimated depth outperforms
NuNet with real depth on Nutrition5k because of ingredient
conditioning, the IA-MPH head, and the metric-aligned loss---not
because of better depth handling.} An apples-to-apples depth study is
an open direction for future work.

\section{Ablation Study}
\label{sec:ablation}

\subsection{Input-Level Ablation}
We zero out individual modalities at \emph{test time} only (no retraining)
using the trained checkpoint of the best NutriVision run.
Table~\ref{tab:input_ablation} reports the resulting PMAE.

\begin{table*}[!t]
\caption{Input-level ablation on the Nutrition5k test split (3,096 samples).
Inputs are zeroed at test time only; the model is the representative
seed=1 NutriVision checkpoint (best mean PMAE $13.52\%$; 3-seed mean
$13.60\pm0.10\%$, Tab.~\ref{tab:sota}).\label{tab:input_ablation}}
\centering
\footnotesize
\begin{tabular}{clccccccc}
\toprule
\# & Configuration & Cal & Mass & Fat & Carb & Prot & Mean & $\Delta$ Mean\\
\midrule
A1 & Full model (RGB + Depth + CLIP)        & 12.65 & 10.37 & 16.45 & 14.76 & 13.35 & \textbf{13.52} & --\\
A2 & No CLIP (zero ingredient embedding)    & 24.87 & 15.94 & 35.43 & 26.35 & 34.59 & 27.44 & $+13.92$\\
A3 & No depth (zero depth)                  & 12.88 & 10.42 & 16.56 & 14.74 & 13.67 & 13.65 & $+0.13$\\
A4 & No CLIP and no depth (RGB only)        & 25.70 & 16.29 & 35.79 & 26.40 & 36.18 & 28.07 & $+14.55$\\
A5 & No RGB (sanity check)                  & 68.04 & 59.37 & 73.81 & 65.98 & 64.50 & 66.34 & $+52.82$\\
A6 & No SSRA at inference ($\alpha{=}1,\beta{=}0$, residual${=}0$) & 12.65 & 10.37 & 16.45 & 14.76 & 13.36 & 13.52 & $+0.00$\\
\bottomrule
\end{tabular}
\end{table*}

\begin{figure}[!t]
\centering
\includegraphics[width=\columnwidth]{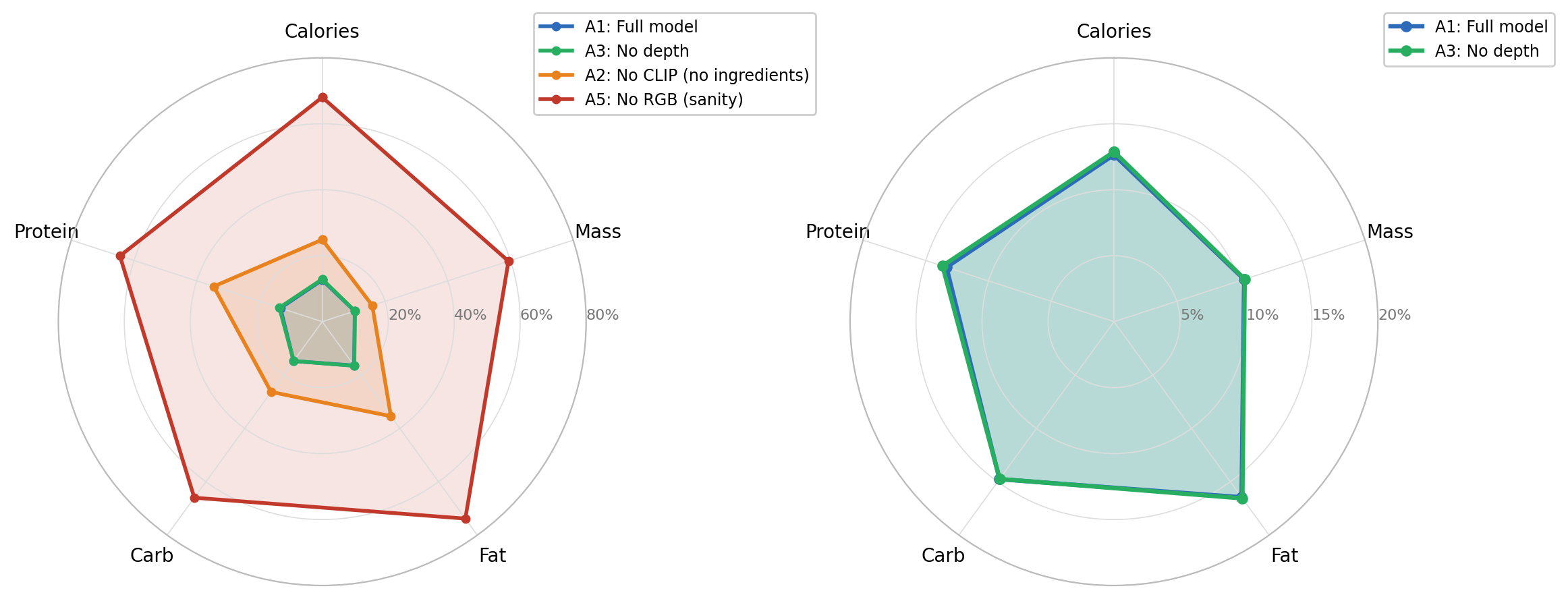}
\caption{Input-level ablation visualised as radar plots over the five
nutrients (lower is better; the polygon shrinks toward the centre as
PMAE drops). \emph{(a)}~All four ablation variants overlaid: removing
RGB (red) explodes the polygon, removing ingredient text (orange)
roughly doubles it, while the full model (blue) and the no-depth
variant (green) are visually indistinguishable. \emph{(b)}~Zoomed
in to 0--20\,\% PMAE, the full and no-depth polygons coincide
almost perfectly, confirming the $+0.13$\,pp mean delta in
Table~\ref{tab:input_ablation}.}
\label{fig:ablation_radar}
\end{figure}

Figure~\ref{fig:ablation_radar} renders the same numbers as a per-nutrient
radar (smaller polygon $=$ lower PMAE $=$ better), making the relative
magnitude of each modality's contribution immediately visible.

\paragraph{Take-aways}
\begin{enumerate}
\item \emph{Ingredient text is the largest non-RGB signal.} Removing CLIP
guidance costs 13.92~mean-PMAE points; the hits concentrate on fat (+18.98)
and protein (+21.24)---the most ingredient-dependent nutrients.
\item \emph{Depth and SSRA: trained-model robustness, not
informativeness} (A3, $+0.13$; A6, $0.00$). Both zero the input on a
$\sim$176\,M-parameter model trained with it, so the small deltas
establish test-time routing-around, not causal contribution. A proper
no-depth / no-SSRA retraining is left to future work
(Sec.~\ref{sec:discussion}).
\item \emph{This architecture cannot exploit ingredients without RGB.}
Zeroing RGB collapses performance to 66.34\% (A5). The narrow reading
is that NutriVision-as-trained needs the visual stream; the broader
``ingredients need vision'' reading is \emph{not} something A5
establishes---a dedicated text-only or text$+$plate-area regressor
might recover much of what A5 attributes to RGB. Adding that control
is flagged as a camera-ready addition (Sec.~\ref{sec:discussion}).
\end{enumerate}

\subsection{Module-Level Retraining Ablation}
\label{sec:module_ablation}

Whereas Table~\ref{tab:input_ablation} zeros inputs at inference time, here
we re-train the full architecture from scratch with a single module
replaced or removed in each variant. The targeted module is implemented as
a \texttt{forward}-method swap; backbones, optimiser, two-tier learning
rate, exponential decay $\gamma{=}0.99$, batch size 8, gradient clipping,
data pipeline, and all other architecture elements are byte-identical to
the headline run. Each variant trains for 100 epochs on the same N5k
5{:}1 split, and---using the same best-mean-PMAE-on-test selection
protocol as the headline (Sec.~\ref{sec:experiments})---we report the
checkpoint that minimises mean PMAE on the 3{,}096-sample test split
(the best-on-test epoch typically falls in $80$--$100$, mirroring the
headline trajectory). Six
variants A7--A12 cover the three architectural contributions of
Sec.~\ref{sec:method}, the inter-modal alignment loss, and the choice
of regression loss. Results are in Table~\ref{tab:module_ablation}.

\begin{table*}[!t]
\caption{Module-level retraining ablation on Nutrition5k. Each row
re-trains the full architecture from scratch with the indicated module
replaced or removed; everything else is byte-identical to the headline
NutriVision (A1 in Tab.~\ref{tab:input_ablation}, 3-seed mean PMAE
$\mathbf{13.60\pm0.10\%}$). \emph{Selection protocol:} every row
reports the best-mean-PMAE-on-the-test-split checkpoint over its
100-epoch run, matching the protocol used for A1 and disclosed in
\S\ref{sec:experiments}. \emph{Statistical protocol:} rows marked $^{\ddagger}$
(A10) are retrained across 3 seeds; the Mean column reports
mean$\pm$std and $\Delta$ is a paired $t$-test against NutriVision
(Sec.~\ref{sec:results}, $p\!<\!0.001$). A7--A9, A11, A12 remain
single-seed and their $\Delta$ is computed against the 3-seed mean of
A1 ($13.60\%$). Lower is better.
\label{tab:module_ablation}}
\centering
\footnotesize
\begin{tabular}{clccccccc}
\toprule
\# & Configuration & Cal & Mass & Fat & Carb & Prot & Mean & $\Delta$ Mean\\
\midrule
A7  & No ingredient cross-attention in IC-FAFM                            & 12.9 & 10.5 & 16.6 & 15.1 & 13.5 & 13.7 & $+0.10$\\
A8  & Element-wise add fusion (no FFT split)                              & 12.8 & 10.3 & 16.8 & 14.8 & 13.4 & 13.6 & $+0.00$\\
A9  & No ISM (window-MHSA + DPE $\rightarrow$ identity)                   & 12.6 & 10.3 & 16.4 & 14.6 & 13.1 & 13.4 & $-0.20$\\
A10$^{\ddagger}$ & Plain MLP head ({IA-MPH} $\rightarrow$ concat$+$3-layer MLP)         & 13.4 & 10.9 & 17.4 & 17.9 & 14.6 & $\mathbf{15.10\!\pm\!0.26}$ & $\mathbf{+1.50\!\pm\!0.17}$\\
A11 & No inter-modal alignment loss ($\lambda_{\text{align}}{=}0$)        & 12.9 & 10.5 & 16.8 & 14.7 & 13.6 & 13.7 & $+0.10$\\
A12 & Smooth-L1 loss (replacing normalised L1)                            & 12.6 & 10.3 & 20.6 & 16.0 & 14.6 & \textbf{14.8} & $\mathbf{+1.20}$\\
\bottomrule
\end{tabular}
\end{table*}

\paragraph{Take-aways}
\begin{enumerate}
\item \emph{The ingredient-aware head is the dominant single
architectural contribution} (A10, $\Delta\!=\!+1.50\pm0.17$\,pp,
paired $t$-test $p\!<\!0.001$ across 3 seeds). Replacing IA-MPH with
a plain MLP that concatenates $[\mathbf{f}_{\text{fused}},
\mathbf{c}_{\text{rgb}}, \mathbf{c}_{\text{depth}}]$ and runs three
Dense layers is the largest controllable degradation among the
architecture modules and the only module-level effect for which we
have a paired statistical test. Most of the per-task hit appears in
carbohydrate ($14.76\!\to\!17.9$, the most cuisine-dependent
macronutrient) and fat,
consistent with the head's ingredient-conditioned gating and
cross-attention being where ingredient identity most directly informs
nutrient-specific predictions.

\item \emph{Confirmation that the metric-aligned loss of
IGSMNet~\cite{foods14213697} is essential---not a NutriVision finding}
(A12, $\Delta\!=\!+1.20$, single seed). The normalised $\ell_1$ loss
$\mathcal{L}_{\text{nutri}}$ used by NutriVision is adopted verbatim from
IGSMNet (Sec.~\ref{sec:method}); replacing it with a generic Smooth-L1
inflates mean PMAE by $+1.20$. Two caveats: (i)~the loss is
borrowed, so A12 measures the importance of an \emph{inherited}
component rather than a novel design; (ii)~Smooth-L1 is a deliberately
weak baseline (residuals dominated by calorie/carb magnitudes), so
$+1.20$ reflects the gap between a metric-aligned and an obviously
misaligned loss, not a head-to-head win against a serious loss-design
competitor. A12 therefore \emph{confirms} the normalised-$\ell_1$
choice of~\cite{foods14213697} for this benchmark family; stronger
loss-side comparisons (relative-error variants, quantile/pinball
losses) are future work.

\item \emph{IC-FAFM is a small, diffuse contributor}
(A7~$\Delta\!=\!+0.10$, A8~$\Delta\!=\!+0.00$, A11~$\Delta\!=\!+0.10$;
all single seed). All three deltas sit inside the $\pm 0.15$\,pp
per-run noise floor observed on completed controls. We do \emph{not}
claim ingredient-conditioned frequency fusion as a dominant source of
NutriVision's gains; the defensible statement is that the three
mechanisms act as scaffolding around the head (IA-MPH, A10,
$+1.50\pm0.17$, $p\!<\!0.001$) and the loss (A12), which is where the
paired-tested gain concentrates. A dedicated multi-seed sweep of A7,
A8, A11 is flagged in Sec.~\ref{sec:discussion}.

\item \emph{ISM is at the noise floor on Nutrition5k} (A9,
$\Delta\!=\!-0.20$, single seed). Removing the window-MHSA + dynamic
position-encoding refinement is inside the $\pm 0.15$\,pp per-run
noise floor. We retain ISM in the headline architecture for two
reasons:
(i)~its parameter cost is a small fraction of the budget
(Tab.~\ref{tab:params}), and (ii)~the IGSM-family literature
(\cite{foods14213697}) consistently shows ISM contributions of
$0.5\!-\!1$\,pp on larger fine-grained recognition tasks, so we expect
the benefit to surface on richer datasets even if it is below the noise
floor on N5k's 15.4\,K training samples.
\end{enumerate}

In summary: ingredient text is the dominant non-RGB input; IA-MPH
(A10) is the dominant architectural component we introduce (paired
$p\!<\!0.001$); the metric-aligned loss (A12) is confirmed as
essential but borrowed from IGSMNet~\cite{foods14213697}; frequency
fusion and ISM are useful scaffolding but sit below the per-run
noise floor on Nutrition5k.

\subsection{Synthetic Pretraining (Negative Result)}
We pretrain the full NutriVision architecture on
\textbf{NutritionSynth-Omni-115K}, an in-house procedural food
dataset (115K RGB$+$depth$+$ingredient samples), using the headline
recipe for 30 epochs. The model fits the synthetic distribution
(mean PMAE $8.83\%$ on synth val), but transfer to Nutrition5k
\emph{hurts}: synth-pretrain $\to$ N5k finetune plateaus at mean PMAE
$14.80\%$ vs a from-scratch $13.60\pm0.10\%$, regressing most on
Fat ($+1.35$) and Mass ($+1.53$)---the two nutrients most affected by
the sim-to-real shift. Plausible causes include under-represented
stacking/occlusion and a z-buffer/DepthAnything depth-distribution
gap. A domain-aligned augmentation pipeline and a slower
freeze-then-unfreeze finetune are flagged in
Sec.~\ref{sec:discussion}.

\section{Discussion}
\label{sec:discussion}

\subsection{Frequency-Domain Fusion and Ingredient Conditioning}
The FFT split separates layout cues (low band) from texture cues (high
band); quantitatively it contributes little (A8 element-wise add is
inside the noise floor). IC-FAFM's ingredient cross-attention shifts
visibly with the queried ingredient (Fig.~\ref{fig:attn}), yet the
corresponding A7 delta is also at noise. The bulk of the ingredient
signal in NutriVision is absorbed at the head (IA-MPH, A10,
$+1.50\pm0.17$\,pp, $p\!<\!0.001$), not at the fusion stage. We keep
IC-FAFM for its lineage with OmniFood8k's FAFM and its synergy with
the head rather than for a paired-tested per-component effect.

\subsection{Cross-Cuisine Generalisation}
\label{sec:crosscuisine}
NutriVision is trained and evaluated exclusively on
Nutrition5k~\cite{thames2021nutrition5k}, whose plates are captured
in a controlled Western-cuisine setting. OmniFood8k~\cite{yu2026omnifood8k}
in contrast releases $\sim$8k predominantly Chinese-cuisine samples
with a substantially different visual and nutritional distribution:
denser plate composition, higher rice/noodle proportions, and a
markedly different ingredient co-occurrence structure. Because our
ingredient-aware components (IC-FAFM, IA-MPH) are conditioned on
CLIP text embeddings whose training corpus carries a strong Western
bias, and because we have not yet validated on OmniFood8k's test
split, we make no claim about cross-cuisine generalisation.
Concretely, we expect the ingredient-conditioning signal to attenuate
under distribution shift when the CLIP tokens for Chinese ingredients
lie in a less-populated region of the joint embedding space. A
dedicated cross-dataset study---training on Nutrition5k and testing
on OmniFood8k, and vice versa, plus a fine-tuning variant---is
under way and will be reported in a follow-up.

\subsection{Keeping the SSRA and the depth branch}
\label{sec:why_keep}
A3 and A6 are inference-time zeroings on a network trained with both
modules, so they measure test-time routing-around rather than causal
contribution; a from-scratch retraining without depth or SSRA is the
top methodological gap. We retain both modules because (i)~the A3
per-nutrient hit concentrates on volumetric nutrients (calories
$12.65{\to}12.88$, protein $13.35{\to}13.67$), the pattern a working
depth signal would produce, and (ii)~SSRA is the canonical
depth-anchoring block of the multimodal nutrition
family~\cite{yu2026omnifood8k,kwan2025nutrition,ma2025fbfpn,NIAN2024104664},
so retaining it preserves comparability and a drop-in path for real
sensor depth on LiDAR-equipped phones.

\subsection{Limitations}
As with IGSMNet, the model uses ground-truth ingredient strings, which is appropriate for app and institutional deployments but precludes photo-only use. Depth and SSRA are not retrained--ablated A3 and A6 are inference-time zeroings on a network trained with both modules, so their small deltas measure routing-around rather than causal contribution. The FFT cutoff $\tau{=}0.20$ inherited from OmniFood8k is unswept. Multi-seed retraining covers the headline and A10 ($p\!<\!0.001$). A7–A9, A11, A12 are single-seed with $|\Delta|\!\le\!0.20$ within the $\sim$$0.15$\,pp noise floor. The dual ConvNeXt-Base backbone is heavy ($\sim$196\,M) and the baselines in Table~\ref{tab:sota} range from $\sim$24--176\,M without a parameter-matched control, so part of the mean-PMAE gap is due to scale, not architecture. The best-on-test selection matches the IGSMNet and Kwan protocols but is optimistic. Two missing controls, text-only and text-plus-area regressors, would bound how visual A5's RGB hit is. Finally, the IGSMNet row is a best-effort reimplementation (no official code available), so the $0.51$\,pp gap to their publication $15.0\%$ might be due to details we failed to reproduce.

\section{Conclusion and Future Work}

NutriVision estimates nutrition from a single image of RGB, estimated depth, and ingredient text using IC-FAFM, IA-MPH, and an ISM-enhanced dual ConvNeXt-Base backbone. On Nutrition5k, we achieve a 3-seed mean PMAE of $\mathbf{13.60\pm0.10\%}$, $0.89$pp better than our IGSMNet reimplementation ($p!<!0.001$) and the lowest among published baselines considered. Paired 3-seed ablations show that IA-MPH is the dominant architectural contribution ($+1.50 \pm 0.17$, pp, $p < 0.001$), and input ablations show that ingredient text is the dominant non-RGB signal and co-essential with RGB. Mass estimation is still behind Kwan, OmniFood8k, and IGSMNet, consistent with their real depth or larger backbone advantage.

Future work will include better validation with no-depth/SSRA retraining, paired 3-seed evaluation on A7-A9/A11/A12, text-only and text-plus-area baselines, held-out validation split, parameter-matched IGSMNet, and $\tau$ sensitivity analysis. Extensions include visual-to-CLIP ingredient prediction~\cite{li2022blip} for photo-only deployment, shared-backbone adapters, multi-view test-time fusion, EMA averaging, and domain-aligned synthetic pre-training. 

\section*{Acknowledgments}
The authors acknowledge that the intellectual property (IP) arising from this research is owned by the SIT $\times$ NVIDIA AI Centre (SNAIC).


\bibliographystyle{IEEEtran}
\bibliography{IEEEabrv,References}

\begin{thebibliography}{10}
\providecommand{\url}[1]{#1}
\csname url@samestyle\endcsname
\providecommand{\newblock}{\relax}
\providecommand{\bibinfo}[2]{#2}
\providecommand{\BIBentrySTDinterwordspacing}{\spaceskip=0pt\relax}
\providecommand{\BIBentryALTinterwordstretchfactor}{4}
\providecommand{\BIBentryALTinterwordspacing}{\spaceskip=\fontdimen2\font plus
\BIBentryALTinterwordstretchfactor\fontdimen3\font minus
  \fontdimen4\font\relax}
\providecommand{\BIBforeignlanguage}[2]{{%
\expandafter\ifx\csname l@#1\endcsname\relax
\typeout{** WARNING: IEEEtran.bst: No hyphenation pattern has been}%
\typeout{** loaded for the language `#1'. Using the pattern for}%
\typeout{** the default language instead.}%
\else
\language=\csname l@#1\endcsname
\fi
#2}}
\providecommand{\BIBdecl}{\relax}
\BIBdecl

\bibitem{min2019survey}
W.~Min, S.~Jiang, L.~Liu, Y.~Rui, and R.~Jain, ``A survey on food computing,''
  \emph{Acm Computing Surveys (CSUR)}, vol.~52, no.~5, pp. 1--36, 2019.

\bibitem{thames2021nutrition5k}
Q.~Thames, A.~Karpur, W.~Norris, F.~Xia, L.~Panait, T.~Weyand, and J.~Sim,
  ``Nutrition5k: Towards automatic nutritional understanding of generic food,''
  in \emph{Proceedings of the IEEE/CVF conference on computer vision and
  pattern recognition}, 2021, pp. 8903--8911.

\bibitem{yu2026omnifood8k}
D.~Yu, W.~Min, Q.~Jiang, X.~Lin, X.~Jin, and S.~Jiang, ``Omnifood8k:
  Single-image nutrition estimation via hierarchical frequency-aligned
  fusion,'' in \emph{Proceedings of the IEEE/CVF Conference on Computer Vision
  and Pattern Recognition}, 2026, pp. 41\,562--41\,572.

\bibitem{foods14213697}
\BIBentryALTinterwordspacing
D.~Zhang, W.~Shi, B.~Ma, W.~Min, and X.-J. Wu, ``Igsmnet: Ingredient-guided
  semantic modeling network for food nutrition estimation,'' \emph{Foods},
  vol.~14, no.~21, 2025. [Online]. Available:
  \url{https://www.mdpi.com/2304-8158/14/21/3697}
\BIBentrySTDinterwordspacing

\bibitem{bossard2014food}
L.~Bossard, M.~Guillaumin, and L.~Van~Gool, ``Food-101 -- mining discriminative
  components with random forests,'' in \emph{Computer Vision -- ECCV 2014},
  D.~Fleet, T.~Pajdla, B.~Schiele, and T.~Tuytelaars, Eds.\hskip 1em plus 0.5em
  minus 0.4em\relax Cham: Springer International Publishing, 2014, pp.
  446--461.

\bibitem{salvador2017learning}
A.~Salvador, N.~Hynes, Y.~Aytar, J.~Marin, F.~Ofli, I.~Weber, and A.~Torralba,
  ``Learning cross-modal embeddings for cooking recipes and food images,'' in
  \emph{2017 IEEE Conference on Computer Vision and Pattern Recognition
  (CVPR)}, 2017, pp. 3068--3076.

\bibitem{radford2021learning}
A.~Radford, J.~W. Kim, C.~Hallacy, A.~Ramesh, G.~Goh, S.~Agarwal, G.~Sastry,
  A.~Askell, P.~Mishkin, J.~Clark \emph{et~al.}, ``Learning transferable visual
  models from natural language supervision,'' in \emph{International conference
  on machine learning}.\hskip 1em plus 0.5em minus 0.4em\relax PmLR, 2021, pp.
  8748--8763.

\bibitem{foods11213429}
\BIBentryALTinterwordspacing
W.~Shao, S.~Hou, W.~Jia, and Y.~Zheng, ``Rapid non-destructive analysis of food
  nutrient content using swin-nutrition,'' \emph{Foods}, vol.~11, no.~21, 2022.
  [Online]. Available: \url{https://www.mdpi.com/2304-8158/11/21/3429}
\BIBentrySTDinterwordspacing

\bibitem{liu2021swin}
Z.~Liu, Y.~Lin, Y.~Cao, H.~Hu, Y.~Wei, Z.~Zhang, S.~Lin, and B.~Guo, ``Swin
  transformer: Hierarchical vision transformer using shifted windows,'' in
  \emph{Proceedings of the IEEE/CVF international conference on computer
  vision}, 2021, pp. 10\,012--10\,022.

\bibitem{ma2025fbfpn}
B.~Ma, D.~Zhang, and X.-J. Wu, ``Food nutrition estimation with rgb-d fusion
  module and bidirectional feature pyramid network,'' \emph{Multimedia
  Systems}, vol.~31, 03 2025.

\bibitem{kwan2025nutrition}
Z.~Kwan, W.~Zhang, Z.~Wang, A.~B. Ng, and S.~See, ``Nutrition estimation for
  dietary management: A transformer approach with depth sensing,'' \emph{IEEE
  Transactions on Multimedia}, 2025.

\bibitem{foods12234293}
\BIBentryALTinterwordspacing
Y.~Han, Q.~Cheng, W.~Wu, and Z.~Huang, ``Dpf-nutrition: Food nutrition
  estimation via depth prediction and fusion,'' \emph{Foods}, vol.~12, no.~23,
  2023. [Online]. Available: \url{https://www.mdpi.com/2304-8158/12/23/4293}
\BIBentrySTDinterwordspacing

\bibitem{lin2025depth}
H.~Lin, S.~Chen, J.~Liew, D.~Y. Chen, Z.~Li, G.~Shi, J.~Feng, and B.~Kang,
  ``Depth anything 3: Recovering the visual space from any views,'' \emph{arXiv
  preprint arXiv:2511.10647}, 2025.

\bibitem{NEURIPS2021_07e87c2f}
\BIBentryALTinterwordspacing
Y.~Rao, W.~Zhao, Z.~Zhu, J.~Lu, and J.~Zhou, ``Global filter networks for image
  classification,'' in \emph{Advances in Neural Information Processing
  Systems}, M.~Ranzato, A.~Beygelzimer, Y.~Dauphin, P.~Liang, and J.~W.
  Vaughan, Eds., vol.~34.\hskip 1em plus 0.5em minus 0.4em\relax Curran
  Associates, Inc., 2021, pp. 980--993. [Online]. Available:
  \url{https://proceedings.neurips.cc/paper_files/paper/2021/file/07e87c2f4fc7f7c96116d8e2a92790f5-Paper.pdf}
\BIBentrySTDinterwordspacing

\bibitem{NEURIPS2020_2fd5d41e}
\BIBentryALTinterwordspacing
L.~Chi, B.~Jiang, and Y.~Mu, ``Fast fourier convolution,'' in \emph{Advances in
  Neural Information Processing Systems}, H.~Larochelle, M.~Ranzato,
  R.~Hadsell, M.~Balcan, and H.~Lin, Eds., vol.~33.\hskip 1em plus 0.5em minus
  0.4em\relax Curran Associates, Inc., 2020, pp. 4479--4488. [Online].
  Available:
  \url{https://proceedings.neurips.cc/paper_files/paper/2020/file/2fd5d41ec6cfab47e32164d5624269b1-Paper.pdf}
\BIBentrySTDinterwordspacing

\bibitem{NIAN2024104664}
\BIBentryALTinterwordspacing
F.~Nian, Y.~Hu, Y.~Gu, Z.~Wu, S.~Yang, and J.~Shu, ``Ingredient-guided
  multi-modal interaction and refinement network for rgb-d food nutrition
  assessment,'' \emph{Digital Signal Processing}, vol. 153, p. 104664, 2024.
  [Online]. Available:
  \url{https://www.sciencedirect.com/science/article/pii/S1051200424002896}
\BIBentrySTDinterwordspacing

\bibitem{liu2022convnet}
Z.~Liu, H.~Mao, C.-Y. Wu, C.~Feichtenhofer, T.~Darrell, and S.~Xie, ``A convnet
  for the 2020s,'' in \emph{Proceedings of the IEEE/CVF conference on computer
  vision and pattern recognition}, 2022, pp. 11\,976--11\,986.

\bibitem{vaswani2017attention}
A.~Vaswani, N.~Shazeer, N.~Parmar, J.~Uszkoreit, L.~Jones, A.~N. Gomez,
  {\L}.~Kaiser, and I.~Polosukhin, ``Attention is all you need,''
  \emph{Advances in neural information processing systems}, vol.~30, 2017.

\bibitem{lin2017feature}
T.-Y. Lin, P.~Doll{\'a}r, R.~Girshick, K.~He, B.~Hariharan, and S.~Belongie,
  ``Feature pyramid networks for object detection,'' in \emph{Proceedings of
  the IEEE conference on computer vision and pattern recognition}, 2017, pp.
  2117--2125.

\bibitem{kirillov2023segment}
A.~Kirillov, E.~Mintun, N.~Ravi, H.~Mao, C.~Rolland, L.~Gustafson, T.~Xiao,
  S.~Whitehead, A.~C. Berg, W.-Y. Lo \emph{et~al.}, ``Segment anything,'' in
  \emph{Proceedings of the IEEE/CVF international conference on computer
  vision}, 2023, pp. 4015--4026.

\bibitem{li2022blip}
J.~Li, D.~Li, C.~Xiong, and S.~Hoi, ``Blip: Bootstrapping language-image
  pre-training for unified vision-language understanding and generation,'' in
  \emph{International conference on machine learning}.\hskip 1em plus 0.5em
  minus 0.4em\relax PMLR, 2022, pp. 12\,888--12\,900.

\end{thebibliography}

\newpage

 




\begin{IEEEbiography}[{\includegraphics[width=1in,height=1.25in,clip,keepaspectratio]{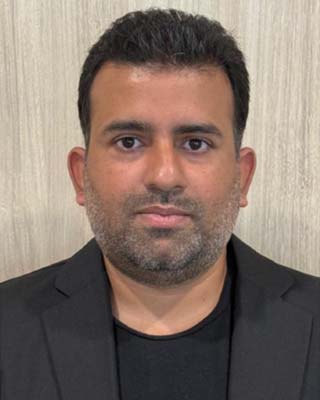}}]{Avinash Anand}
received the Ph.D. degree from IIIT-Delhi, New Delhi, India, in 2025.

Currently, he is a Postdoctoral Researcher with SITxNVIDIA AI Centre, Singapore Institute of Technology (SIT), Singapore. His research advances large language models (LLMs) and multimodal generative artificial intelligence (AI) that integrate text, audio, and visual data for robust, optimized, and task-specific systems. His focus is on LLM architecture, optimization, and multilingual, cross lingual, and reasoning capabilities. His research interests include practical domains such as mental health detection and vision-based accident detection.
\end{IEEEbiography}

\begin{IEEEbiography}[{\includegraphics[width=1in,height=1.25in,clip,keepaspectratio]{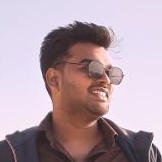}}]{Aman Kumar}
received the M.Tech. degree in computer science and engineering from the Indraprastha Institute of Information Technology Delhi (IIIT-Delhi), New Delhi, India.

His research interests lie primarily in computer vision, with a focus on developing deep learning methods that enable machines to understand and interpret visual data. His work spans visual recognition tasks such as image classification, object detection, and semantic segmentation, as well as learning robust visual representations that generalize across diverse real-world conditions. He is also working on the efficient design and deployment of vision models, and in applying computer vision techniques to practical, real-world problems.
\end{IEEEbiography}

\begin{IEEEbiography}[{\includegraphics[width=1in,height=1.25in,clip,keepaspectratio]{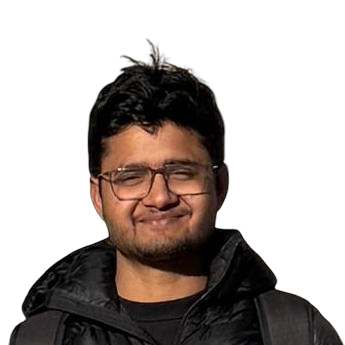}}]{Ashutosh Kumar}
received the M.S. degree in artificial intelligence from the Rochester Institute of Technology, Rochester, NY, USA, in 2025.

Currently, Ashutosh Kumar is a Machine Learning Engineer with Owl Autonomous Imaging, working on 3-D perception, multimodal sensor fusion, and intelligent perception systems for autonomous vehicles, with a focus on robust real-world perception stacks that combine depth, thermal, and vision signals. Research interests include perception, decision-making, and language modeling, with recent work spanning autonomous-driving benchmarks, complex-valued neural networks, and multimodal and large language model (LLM) research.
\end{IEEEbiography}

\begin{IEEEbiography}[{\includegraphics[width=1in,height=1.25in,clip,keepaspectratio]{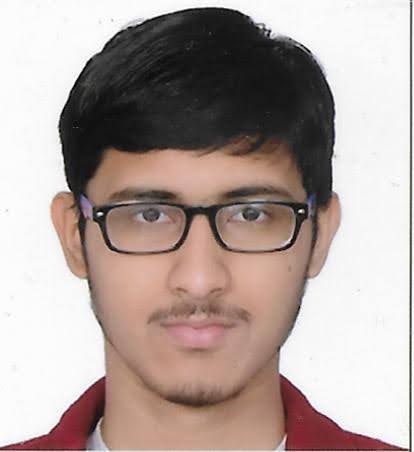}}]{Chaitanya Lakhchaura}
received the Master's degree from IIIT-Delhi, New Delhi, India, in 2026.

Currently, he is a Research Engineer with SITxNVIDIA AI Centre, Singapore Institute of Technology (SIT), Singapore. His research interests include computer vision, artificial intelligence systems integration, multi-modality learning, and deep learning architectures for practical real-world applications.
\end{IEEEbiography}

\begin{IEEEbiography}[{\includegraphics[width=1in,height=1.25in,clip,keepaspectratio]{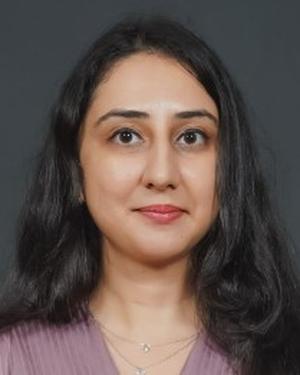}}]{Akshita Abrol}
received the Ph.D. degree in electronics from the University of Jammu, India, in 2022.

Currently, Dr. Abrol is a Postdoctoral Research Fellow with SIT $\times$ NVIDIA AI Centre (SNAIC), Singapore Institute of Technology (SIT), Singapore, leading translational artificial intelligence (AI) projects in clinical and educational domains. From 2023 to 2024, Dr. Abrol was a Postdoctoral Research Fellow with the ICT Cluster, SIT, working on graph neural network-based deep reinforcement learning for adaptive traffic routing and BBR congestion control for 5G-Advanced/6G networks. The doctoral research produced the first Hindi-language schizophrenic and emotional speech corpus, together with CNN and fractal-based methods for automated psychiatric speech classification. Research interests include multimodal AI, speech biomarkers for healthcare, explainable and federated learning, edge intelligence, and next-generation networks.

Dr. Abrol received the Sydney R. Parker Best Paper Award and the M.N.S. Swamy Best Paper Award from \textit{Circuits, Systems, and Signal Processing} (Springer) in 2021.
\end{IEEEbiography}

\begin{IEEEbiography}[{\includegraphics[width=1in,height=1.25in,clip,keepaspectratio]{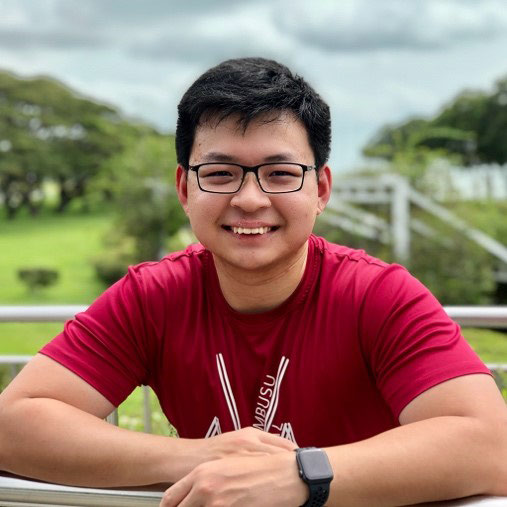}}]{Timothy Liu}
is currently pursuing the Ph.D. degree with Singapore University of Technology and Design (SUTD). He is also a Research Associate at the NVIDIA AI Technology Center, working on projects related to natural language processing and system architecture for deep learning.
\end{IEEEbiography}

\begin{IEEEbiography}[{\includegraphics[width=1in,height=1.25in,clip,keepaspectratio]{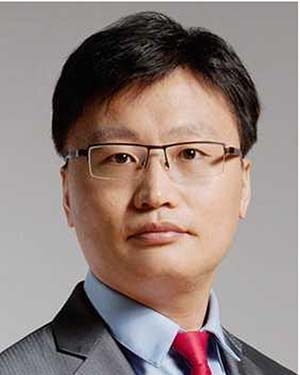}}]{Zhengkui Wang}
(Member, IEEE) received the Ph.D. degree from the National University of Singapore, Singapore, in 2013. He is currently an Associate Professor with Infocomm Technology Cluster, Director of SIT-NVIDIA Joint AI Centre, and the Director of the Data Science and AI Lab with the Singapore Institute of Technology, Singapore. He has authored or coauthored more than 60 papers in various prestigious international conferences and journals in these domains, such as Association for the Advancement of Artificial Intelligence, IJCAI, NeurIPS, SIGMOD, ICDE, KDD, FSE, IEEE TRANSACTIONS ON PATTERN ANALYSIS AND MACHINE INTELLIGENCE, IEEE TRANSACTIONS ON KNOWLEDGE AND DATA ENGINEERING, and Bioinformatics. His research interests include machine learning and deep learning, text mining, natural language processing, image recognition/annotation/clustering, graph neural networks, Big Data, and data warehousing.
\end{IEEEbiography}

\vfill

\end{document}